Ю.В. Паржин

*Национальный технический университет «ХПИ», Харьков*

# ДЕТЕКТОРНЫЙ ПРИНЦИП ПОСТРОЕНИЯ ИСКУССТВЕННЫХ НЕЙРОННЫХ СЕТЕЙ КАК АЛЬТЕРНАТИВА КОННЕКЦИОНИСТСКОЙ ПАРАДИГМЕ

*Искусственные нейронные сети (ИНС) являются неадекватными биологическим нейронным сетям. Эта неадекватность проявляется в использовании устаревшей модели нейрона и коннекционистской парадигмы построения ИНС. Результатом данной неадекватности является существование множества недостатков ИНС и проблем их практической реализации. В статье предлагается альтернативный принцип построения ИНС. Этот принцип получил название детекторного принципа. Основой детекторного принципа является рассмотрение свойства связности входных сигналов нейрона. В данном принципе используется новая модель нейрона-детектора, новый подход к обучению ИНС – встречное обучение и новый подход к формированию архитектуры ИНС.*

## Введение

Коннекционистская парадигма в настоящее время является единственной моделью построения искусственных нейронных сетей (ИНС), которая имеет практическое применение. Данная парадигма объединяет взгляды исследователей на: принципы формирования и модели архитектуры сетей, модели нейроэлементов, роль весовых коэффициентов синаптических связей, принципы формирования реакции нейроэлементов, процедуры обучения ИНС. Однако коннекционистские ИНС являются неадекватными биологическим нейросетям. Это определяет множество известных проблем данных ИНС [1, 2], среди которых:

- проблемы построения эффективных алгоритмов обучения;
- проблема "переобучение" сети (overfitting);
- неуправляемое выделение только локальных признаков распознавания (что особенно ярко проявляется в сверточных ИНС (convolutional neural network (CNN));
- проблема стабильности-пластичности памяти;
- проблема классификации-идентификации в процессе распознавания;
- проблема размерности сети и объема памяти;
- проблема реализации ассоциативного распознавания;
- проблема инвариантности к аффинным преобразованиям и деформационным искажениям распознаваемых образов;
- отсутствие универсальности (ограниченность круга решаемых задач каждой отдельной ИНС);
- отсутствие связи коннекционистского и символьного подходов, что делает невозможным генерирование (вывод) в ИНС новых знаний, т.е. их использование для интеллектуального анализа и принятия решений и др.

Большинство из этих проблем имеют концептуальный характер, с некоторыми проблемами борются с помощью специальных архитектурных решений и алгоритмов обучения [3], однако в рамках коннекционистской парадигмы не существует общего подхода, устраняющего все проблемы. В этих условиях проектирование нейросети для решения практической задачи является в большой степени искусством.

Неадекватность коннекционистских ИНС биологическим прототипам проявляется:

**1.** В неадекватности используемых моделей нейронов. Современные коннекционистские ИНС в практических приложениях, как правило, используют формальную модель нейрона - модель МакКаллока-Питтса (МКП) либо ее модификации и разновидности: сигмоидальный нейрон; нейрон типа "Адалайн"; Паде-нейрон; нейрон с квадратичным сумматором; сигма-пи нейрон; нейроны Гроссберга: "instar", "outstar", слоя распознавания сети АРТ; радиальный нейрон; нейрон Хебба; нейрон Фукушимы; стохастические нейроны; векторные нейроны; импульсные (спайковые) нейроны и др. [2, 3]. Модель МКП была создана еще в 1943 году, и она не отражает современные взгляды нейрофизиологии и нейропсихологии: на роль дендритного дерева и его отдельных структур, в том числе и синапсов, в процессе интеграции возбуждающих постсинаптических потенциалов (ВПСП) [4]; на значение и многообразие форм реакций биологического нейрона [5]; на процесс интеграции ВПСП и формирование реакции биологического нейрона [6].

**2.** В неадекватности применяемых моделей обучения: "с учителем" (supervised learning), "без учителя" (unsupervised learning); с подкреплением (reinforcement learning); а также большинства применяемых методов (процедур) обучения ИНС, например, широко используемого метода обратного распространения ошибки (error back propagation) и др. Биологическая неправдоподобность данных моделей и методов обучения, кроме их вычислитель-ной сложности, прежде всего, заключается в том, что процесс обучения отдельных нейронов в мозге не ведет к переобучению других нейронов или всей сети в целом. Биологический нейрон (очевидно, что в данном случае имеет смысл говорить только корковых, например, пирамидальных нейронах) может обучиться только на одном единичном примере и, уж во всяком случае, ему не нужно множество эпох обучения, необходимых для обучения большинства коннекционистских ИНС. Обучение отдельных биологических нейронов и обучение сети биологических нейронов (нейронных модулей) - это разные процессы, или точнее - разные стороны общего процесса обучения мозга, обусловленного как экзогенными, так и эндогенными факторами. Кроме того, очевидно, что, в отличие от коннекционистских парадигм обучения, обучение нейронов и нейронных модулей мозга это рациональная процедура, в результате которой формируется только необходимое и достаточное для решения конкретной задачи количество синаптических связей и обученных нейронов [7]. Это не противоречит весьма высокой надежности биологической нейронной системы, которая достигается многими механизмами резервирования (например, дублирования) и адаптации [8].

**3.** В неадекватности архитектур ИНС. Практически используемые коннекционистские ИНС имеют, как правило, многослойную и зачастую неоднородную архитектуру. Так, например, глубокие нейронные сетей (англ. – "deep neural network" (DNN)) могут иметь десятки и даже сотни слоев [9]. Нейроны каждого внутреннего слоя определяют шаг в обработке информации, заключающийся в формировании признаков распознавания определенной степени общности (определенного уровня интеграции). Характер межнейронных связей между всеми слоями либо между слоями определенного вида, как правило, однотипный.

Безусловно, что сложнейшая структура связей между нейронами мозга строится на протяжении всей жизни организма, и осуществить детальное моделирование данного процесса в настоящее время, очевидно, невозможно. Однако если рассмотреть зрительную систему, которая является объектом моделирования многих ИНС, то нам известно, что:

- для распознавания цельных изолированных зрительных образов, например: лиц, людей, животных, цветов, геометрических фигур, символов и др., вероятно, необходимо всего два этапа или два шага обработки входной информации (не считая этапа сенсорного восприятия): первичной обработки информации, которая происходит в первичной (стриарной) зрительной коре мозга V1, и вторичной обработки - во вторичной зрительной коре мозга V2. В первичной зрительной коре выделяются (детектируются) простые или непроизводные признаки распознавания отдельных элементов образа, например: отрезки прямых линий определенной длины и ориентации, точки концов отрезков, точки пересечения отрезков, угловые точки и др. [10]; а во вторичной - сложные цельные образы [11];

- архитектура первичной зрительной коры имеет проекционный или топический (ретинотопический) характер, однако архитектура вторичной зрительной коры уже не имеет этой организации [11, 12];

- в зрительной системе существует раздельное формирование взаимосвязанных информационных потоков: дорсального (заднего) и вентрального (переднего), берущих начало в первичной зрительной коре и объединяющихся, в частности, во вторичной зрительной коре. По дорсальному потоку передается информация о местоположении стимулов в рецепторном поле восприятия, а по вентральному - информация о самом стимуле, необходимая для его распознавания [13, 14]. Эти информационные потоки образуют системы "Где" и "Что" зрительного анализатора.

**4.** В неадекватности моделируемых процессов обработки информации. Процесс восприятия и передачи информации между слоями в большинстве коннекционистских ИНС является параллельным на всем пространстве рецепторного поля. Однако в мозге только на первичном уровне обработки зрительной информации осуществляется параллельное и преаттентивное (неосознанное) детектирование первичных признаков распознавания [15]. Кроме того, параллельное восприятие и последующая обработка информации на первичном уровне происходит не на всем пространстве рецепторного поля, а лишь на выделенном системой внимания сегменте зрительной сцены [16]. На вторичном уровне обработка информации во многом определяется последовательным процессом, связанным с реализацией активного перцептивного акта [17].

## Результаты исследований

Одна из основных причин неадекватности коннекционистских ИНС заключается в существовании свойства связанности (англ. – "connectivity") весовых коэффициентов синаптических связей нейронов. Это свойство проявляется в существовании зависимости (причинно-следственной связи) изменения весовых коэффициентов одного нейрона от изменения весовых коэффициентов другого нейрона в процессе обучения. Решение практически любой задачи распознавания образов в коннекционистских ИНС осуществляется в результате целенаправленной модификации связанных весовых коэффициентов.

Однако в биологическом нейроне "сила" синапса, моделью которой является весовой коэффициент, не зависит от "силы" синапсов других нейронов. "Сила" биологического синапса проявляется в его способности модулировать входной сигнал и влияет на величину локального ВПСП [4, 6]. Понятие "силы" синапса связывают с различными формами его пластичности [6]. Однако величина локального ВПСП в основном определяется: амплитудой или частотой входного сигнала, временными параметрами активизации (возбуждения) данной синаптической связи, а также более тонкими механизмами, например, влиянием сложного взаимодействия тормозных и возбуждающих синапсов, находящихся в ближайшем окружении данной синаптической связи [18].

Также вероятно, что величина ВПСП, а значит и "сила" синапсов не является определяющим фактором в возбуждении биологического нейрона - формировании потенциала действия. Это связано с тем, что как бы ни был велик ВПСП, генерируемый, например, на конце дендритной ветки, в силу кабельных свойств пассивного дендрита (кабельная теория Ролла) он может уменьшиться в 100 раз и иметь задержку от 5 до 50 мсек [19]. Тем не менее, учитывая морфологические особенности дендрита, удаленные от сомы синапсы могут играть весьма существенную роль в процессе возбуждения нейрона [20].

Пассивные свойства дендрита позволяют осуществлять нелинейную интеграцию локальных ВПСП. Таким образом, роль синаптической взаимосвязи в возбуждении нейрона определяется не только биохимическими процессами, происходящими в синапсе, и морфологией синапса, но также и мембранными свойствами и морфологией дендритного дерева.

Кроме того, существование активной дендритной проводимости, связанной с функционированием потенциал-зависимых ионных каналов дендрита, полностью меняет картину интеграции ВПСП: ВПСП могут ускоряться либо тормозиться; усиливаться либо затухать; суммироваться линейно, сублинейно либо суперлинейно; менять амплитуду и временные характеристики; регенерироваться и проявлять другие формы модификации [21 - 24]. Это говорит о роли дендритного дерева как сложного фильтра входной информации, где именно морфология дендритного дерева определяет условия и параметры фильтрации.

Следовательно, роль эндогенных факторов в процессе интеграции ВПСП является решающей, что заставляет пересмотреть определение роли весового коэффициента синапса в модели нейрона, а также осуществить моделирование информационных интегративных процессов, происходящих в дендритном дереве. Можно сделать вывод, что роль синапса и генерируемого ним ВПСП в возбуждении нейрона зависит от его местоположения на дендритном дереве и его взаимосвязи (взаимодействия) с другими синапсами.

**1. Связность входных сигналов. Основы детекторного подхода.** Введем понятие связности (англ. – "binding") входных сигналов и весовых синаптических коэффициентов одного нейрона. Дадим следующее определение.

*Определение 1.* Под связностью сигналов во входном векторе и, следовательно, весовых синаптических коэффициентов одного нейрона будем понимать их взаимозависимость, ведущую к возбуждению нейрона. Эта взаимозависимость моделирует морфологические и интеграционные свойства дендритного дерева. Основными параметрами связности являются:

1) абсолютное (по отношению к соме) и относительное (по отношению к другим близлежащим синапсам) местоположение синаптических связей на его дендритном дереве - пространственная связность;

2) время поступления входных сигналов - временная связность.

*Определение 2.* Назовем пространственную связность элементов входного вектора сигналов нейрона, зависящую от строения (морфологии) и свойств дендритного дерева и ведущую к возбуждению нейрона - связностью на дендритном дереве или относительной пространственной связностью.

Таким образом, связность - это свойство, которое качественно характеризует целостность входного вектора сигналов как независимой единицы обработки информации и обучения единичного нейрона. Именно установление связности элементов входного вектора сигналов определяет параметры фильтрации входящей в нейрон информации. Свойство связности меняет смысл и значение весового коэффициента, а также самой процедуры обучения.

Временная связность определяется синхронностью (когерентностью) поступления элементов вектора входных сигналов и имеет экспериментальные подтверждения в нейрофизиологических исследованиях [25]. Данный вид связности отражает один из аспектов решения нейропсихологической проблемы связности ВР ("binding problem") [26 ].

Пространственная связность отражает еще один аспект этой проблемы. Однако моделирование относительной пространственной связности синаптических контактов на дендритном дереве связано с проблемой моделирования не только структуры и свойств дендритного дерева каждого нейрона, но и специфической для каждого нейрона структуры межнейронных связей. Такой подход к моделированию пространственной связности в настоящее время является сложнейшей и нерешенной проблемой.

Решением проблемы моделирования пространственной связности при построении ИНС является переход от рассмотрения относительной пространственной связности к абсолютной пространственной связности.

*Определение 3.* Под абсолютной пространственной связностью будем понимать взаимосвязь элементов входного вектора сигналов нейрона ИНС, а именно абсолютных адресов пресинаптических нейронов, на основе общих значений пространственных индексов, ведущую к возбуждению нейрона.

Это значит, что каждый нейрон в модуле обработки информации ИНС, или в целом в ИНС, должен иметь собственный уникальный адрес (номер). Этот адрес нейрон передает как составляющую своей реакции всем постсинаптическим нейронам. Про-странственная связность адресных составляющих вектора входных сигналов постсинаптического нейрона определяется на основе общих значений других составляющих реакций пресинаптических нейронов - пространственных индексов. Абсолютная пространственная связность, безусловно, не является биологически правдоподобной. Однако такой подход позволяет осуществить моделирование биологически мотивированной относительной пространственной связности.

Таким образом, нейрон ИНС возбуждается, если на его вход подается вектор пространственно связных сигналов, образующий определенный признак распознавания, который задан структурно (на первичном уровне обработки информации) либо которому обучен данный нейрон (на вторичном уровне). Адресная составляющая реакции возбужденного нейрона нумерует данный признак распознавания для последующей обработки, а пространственная составляющая реакции служит основой для определения связности с другими реакциями на входе постсинаптического нейрона. Данный подход является одним из возможных решений известной нейропсихологической проблемы "нейронного кода" [27].

Тогда, каждый подобный нейрон является детектором определенного признака, а ИНС, состоящую из нейронов-детекторов, можно назвать детекторной ИНС (ДИНС). Детекторный принцип построения ИНС определяет: формализацию процесса обработки информации в ДИНС; информационную модель нейрона-детектора и других нейронов; метод обучения ДИНС; архитектуру ДИНС и правила ее построения.

Рассмотрим ДИНС для распознавания объектов "Контурного мира" (КМ) - плоских, статических, изолированных (не соединенных) контурных изображений, состоящих из отрезков прямых, где "1" кодирует точку контура изображения единичной толщины, а "0" - фон. К объектам КМ, например, относятся: геометрические фигуры на плоскости, знаки или символы алфавита, цифры и т.д.

Система пространственной индексации моделирует структуру пересекающихся рецептивных полей (РП) нейронов в поле зрительного восприятия и является основой для моделирования системы "Где" зрительного анализатора. Синхронное возбуждение нейронов-детекторов признаков распознавания системы "Что" и нейронов-детекторов местоположения стимулов (пространственных нейронов-детекторов) в рецепторном поле системы "Где", определяет пространственно-временную связность реакций этих нейронов на всех уровнях обработки информации. Тогда, пространственный индекс *I*, представляющий собой номер (адресную составляющую реакции) возбужденного пространственного нейрона-детектора, включается (инкапсулируется) в реакцию синхронно возбужденного нейрона-детектора признаков распознавания (рис. 1).

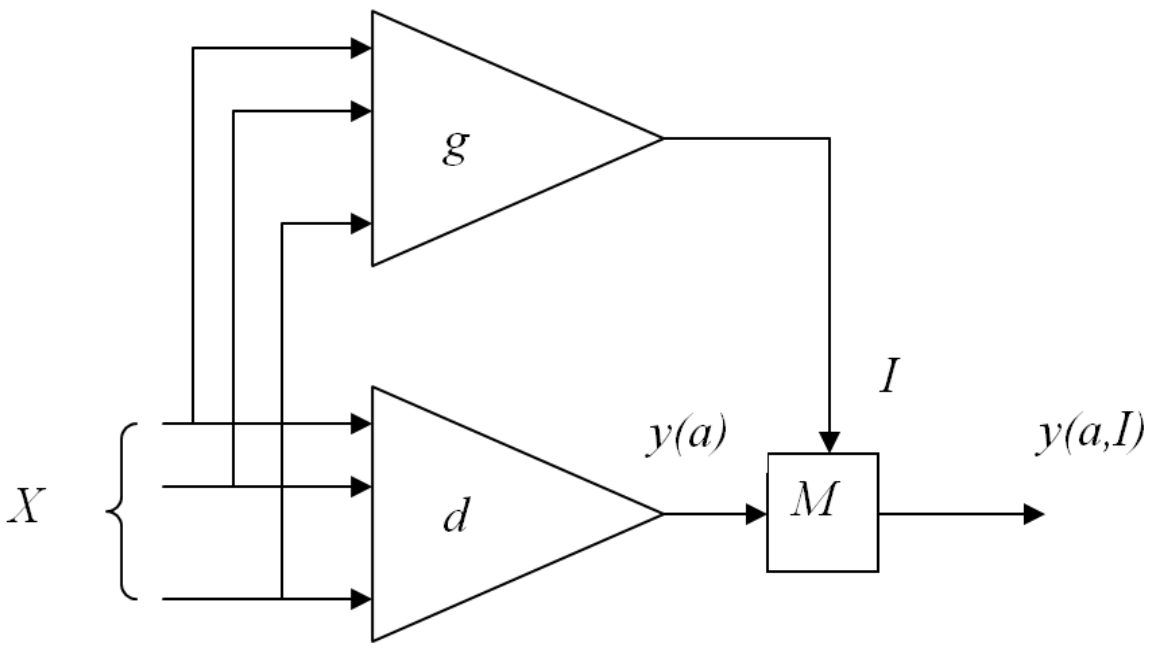

Рис. 1. Схема инкапсуляции пространственного индекса I в реакцию нейрона-детектора

На рис. 1 пространственный нейрон-детектор g системы "Где" и нейрон-детектор признаков распознавания d системы "Что" связаны ретинотопической проекцией с сенсорами системы восприятия, от которой на входы данных детекторов синхронно подаются сигналы X, воспринимаемого стимула. Детектор g формирует реакцию I – пространственный индекс, соответствующий местоположению РП, в который попадает данный стимул. В блоке М детектора d осуществляется инкапсуляция реакции I в реакцию y(a), где a - адресная составляющая реакции, и формирование реакции y(a,I).

Для определения структуры и значений пространственного индекса I, рассмотрим наиболее простой и, вероятно, не самый биологически правдоподобный подход к построению и взаимосвязи рецептивных полей [10].

Представим модель рецепторного поля восприятия в виде прямоугольной матрицы рецепторов восприятия М размерностью l×l элементарных структурных единиц распознаваемого образа - точек, где местоположение каждой точки p определяется номером строки i и номером столбца j матрицы. Обозначим пространственный индекс точки p как $p_{ij}$.

Далее, зададим минимальное рецептивное поле – окно восприятия W минимальной размерности n×n на матрице восприятия M (рис. 2). Сдвиг окна W1 на один столбец вправо (рис. 2) или вниз за n-1 шаг формирует соответствующие области WTA - конкуренции нейронов, рецептивные поля которых попадают в пересекающиеся окна. Области конкуренции - сегменты рецепторного поля M, в которых, в зависимости от максимального попадания стимула в одно окно, в результате конкуренции нейронов-детекторов системы "Где" определяется нейрон-победитель с максимальным уровнем реакции. Адресная составляющая реакции победившего нейрона и будет определять пространственный индекс местоположения данного стимула. Очевидно, что для реализации процедуры конкурентного разделения пространственных индексов необходимо, чтобы пространственно связные стимулы лежали в двух последовательно расположенных и взаимно непересекающихся окнах (в общем случае может быть разной размерности), т.е. находились в области связности (пример на рис. 2).

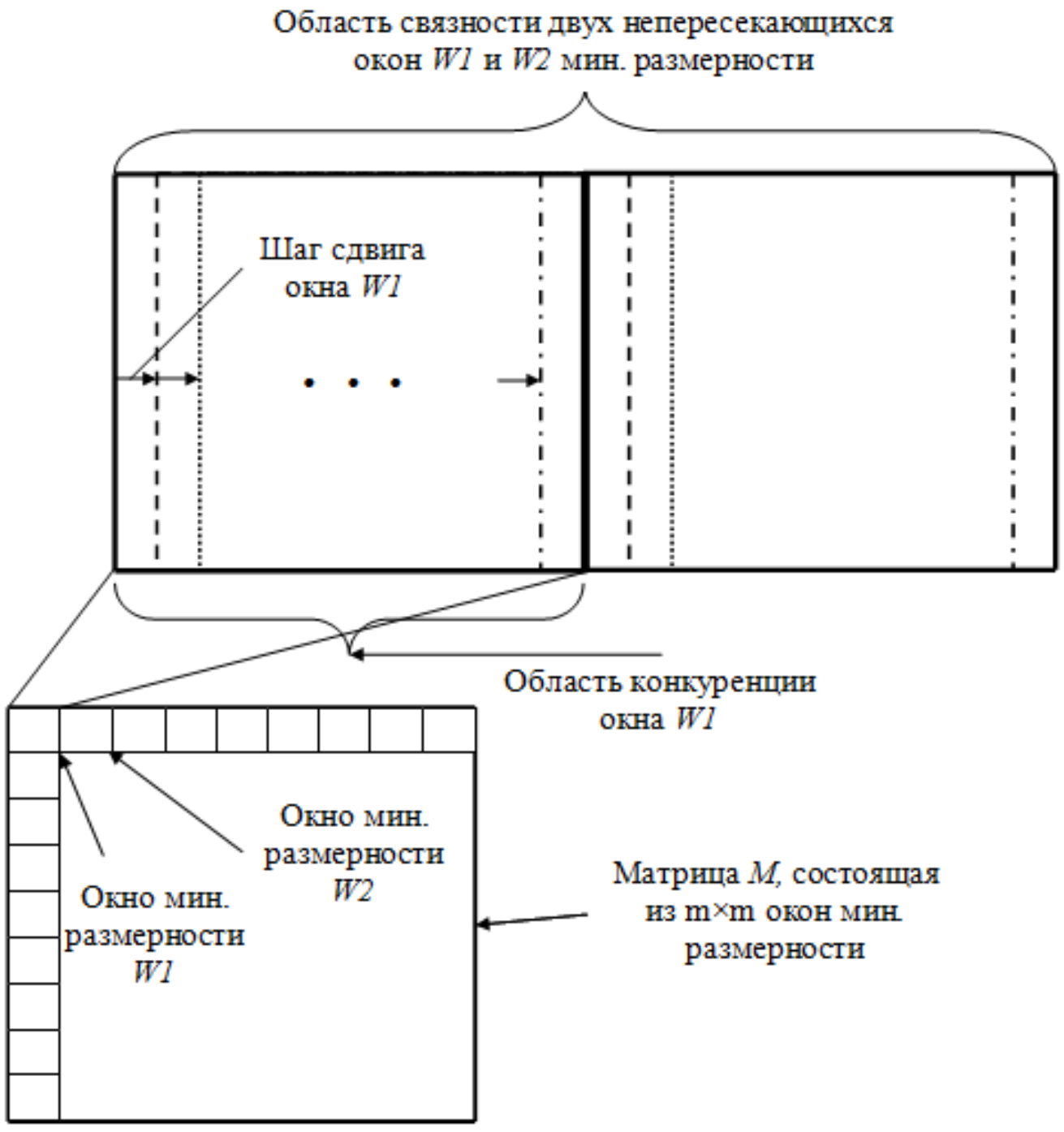

Рис. 2. Структура окон восприятия W на матрице М и схема формирования области конкуренции и связности окон

Таким образом, точка соединения структурных элементов, например, отрезков, должна находиться на общей границе окон, каждое из которых является РП нейрона-победителя в своей конкурентной области. Для приведенной на рис. 2 структуры окон, это возможно для всех окон за исключением тех, которые лежат на границе матрицы М. Для решения этой проблемы также введем сдвиг на n-1 шаг крайних левых и крайних верхних окон в поле восприятия соответственно влево и вверх, что позволит сформировать "усеченные" окна. Эти окна будут формировать области связности слева и сверху для множества окон, которые лежат в области конкуренции пограничных окон.

Таким образом формируется 1-й уровень иерархической структуры пространственных окон минимальной размерности в поле восприятия (в качестве нулевого уровня будем рассматривать местоположение точек в матрице М). Каждое окно на данном уровне будет иметь свой "адрес" или пространственный индекс, который назовем индексом пространственного окна и представим как: $W^1_{j\to k}$, где: 1- первый уровень иерархии; j – номер окна в структуре матрицы М размерностью m×m минимальных окон; → (либо другие стрелки: вниз, вверх или влево для соответствующих окон) - направление смещения окна j на величину k, принимающую на данном уровне значение от 1 до n-1.

На последующих уровнях иерархической структуры пространственных окон будем последовательно увеличивать размерность окна с (n+1)×(n+1) элементов на 2-м уровне до окна размерностью l×l элементов, равному по величине размерности матрицы М на m-м уровне иерархии. На каждом уровне иерархии применяется такая же процедура сдвига окон, как и на 1-м уровне. Тогда:

1) пространственный индекс $I(x_i)$ отрезка $x_i$ будет иметь следующую структуру:

$$I(x_i)=(p1_{ij}(x_i),\ p2_{ij}(x_i),\ W^i_{j*k}(x_i)), \qquad (1)$$

где: $p1_{ij}$ и $p2_{ij}$ – пространственные индексы точек p1 и p2 – концов отрезка $x_i$; $W^i_{j*k}$ - индекс пространственного окна, i – уровень пространственной иерархии окна, * – соответствующее направление сдвига базового окна j на k шагов;

2) пространственные индексы точек концов отрезка являются связными по значению одного пространственного окна $W^i_{j*k}$:

$$I(p1(x_i))=(p1_{ij}(x_i),\ W^i_{j*k}(x_i)), \qquad (2)$$

$$I(p2(x_i))=(p2_{ij}(x_i),\ W^i_{j*k}(x_i)), \qquad (3)$$

где $p1(x_i)$ – первая конечная точка отрезка $x_i$; $p2(x_i)$ – вторая конечная точка отрезка $x_i$.

3) пространственный индекс $I(x_j)$ угла $x_j$:

$$I(x_j)=(p_{ij}(x_j),\ W^i_{j*k}(x_j)) \qquad (4)$$

где $W^i_{j*k}(x_j)$ – является окном минимальной размерности $W^i_{min(j*k)}(x_j)$ в котором находится вершина угла - особая точка (отличная от точки конца отрезка) с пространственным индексом $p_{ij}(x_j)$.

*Определение 4*. Элемент $x_i$ входного вектора сигналов нейрона-детектора, связный с другими элементами в процессе обучения, назовем модой $mod(x_i)$.

Моды, которые несут информацию о типе структурного элемента изображения (например: точки конца отрезка, угла, отрезка, либо всем изображении в целом), являются структурными модами $mod(s_i)$. Соответственно, нейроны-детекторы, реакции которых формируют структурные моды, являются структурными.

На первичном уровне обработки информации ДИНС детектируются перечисленные выше непроизводные структурные элементы распознаваемого образа, а на вторичном уровне - производные структуры - цельные образы.

Моды, которые несут информацию о типе и значении первичной (данной в восприятии) или непроизводной характеристики, детектируемой на первичном уровне обработки информации ДИНС (например: длине отрезка, ориентации отрезка, величине угла), являются непроизводными характеристическими модами $mod(h_{ij})$, а формирующие их нейроны-детекторы являются характеристическими. На вторичном уровне обработки информации детектируются характеристики производные от первичных характеристик. Этот процесс и его значение будут рассмотрены ниже.

Так как процесс детектирования признаков распознавания: структурных элементов и их характеристик на первичном уровне обработки информации (в первичной зрительной коре) достаточно хорошо изучен [10], детально рассмотрим только процесс обработки информации на вторичном уровне. Принцип абсолютной адресации позволяет уже на этом уровне отказаться как от полносвязной, так и от ретинотопической структуры межнейронных связей и использовать связь между нейронами первичного и вторичного уровней типа "общая шина".

Для моделирования связности мод в структуре дендритного дерева введем следующие информационные элементы и их структуры.

*Определение 5.* Каждая структурная мода $mod(s_i)$ является связной с одной или несколькими соответствующими характеристическими модами $mod(h_{ij})$ в структуре вектора входных сигналов нейрона-детектора. Данную структуру назовем модальной группой $Mg(s_i)$ структурной моды $mod(s_i)$.

$$Mg(s_i)=<mod(s_i), mod(h_{i1}), mod(h_{i2}), \ldots, mod(h_{in})>. \quad (5)$$

Связность структурной моды $mod(s_i)$ с характеристическими модами $mod(h_{ij})$ $(j=1,n)$ в модальной группе $Mg(s_i)$ представлена в виде неориентированного графа на рис. 3.

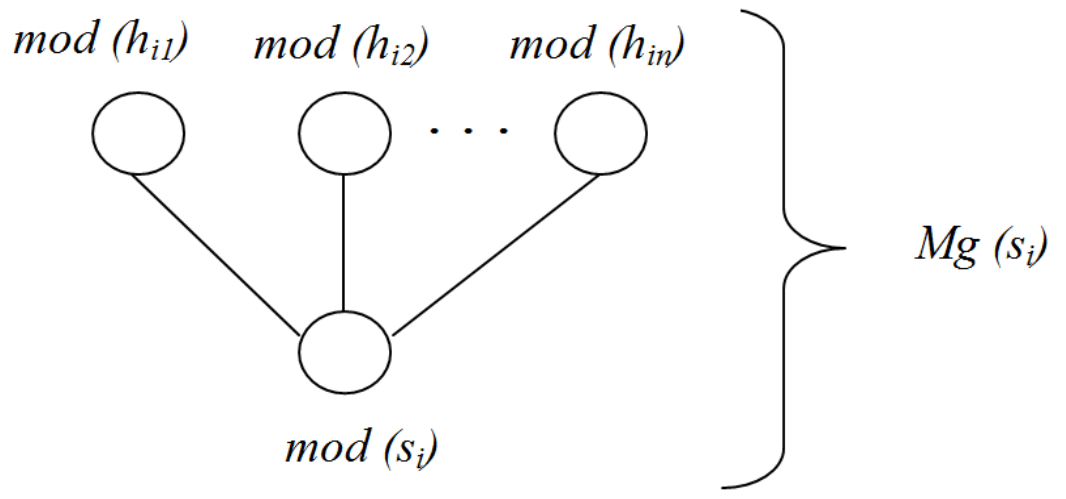

Рис. 3. Модальная группа $Mg(s_i)$ структурной моды $mod(s_i)$

Модальная группа является единицей, а мода – субъединицей информации, обрабатываемой в нейроне-детекторе.

Пространственная связность характеристических мод со структурной модой в модальной группе входного вектора сигналов нейрона-детектора ДИНС осуществляется на основе совпадения значений пространственных индексов реакций соответствующих пресинаптических характеристических нейронов-детекторов и пресинаптического структурного нейрона-детектора.

Таким образом, можно говорить о связности всех мод в модальной группе и связности модальных групп в векторе входных сигналов на основе пространственной связности их структурных мод.

*Определение 6.* Назовем концептом $Con(d_k)$ (от лат. conceptus - понятие; в лингвистике - содержание понятия, смысловое значение) нейрона-детектора $d_k$ минимизированный в процессе обучения (в моменты времени обучения tr) связный граф $G_k$, вершинами которого являются структурные и характеристические моды, а ребрами – отношения пространственной связности, определяемые пространственными индексами I.

$$Con(d_k)=min_{tr}(G_k), \quad (6)$$

где $$G_k := (M, E), \quad (7)$$

M – множество мод (вершин графа); E – множество ребер графа $G_k$ (множество отношений пространственной связности).

Таким образом, концепт $Con(d_k)$, сформированный в процессе обучения, образует память нейрона-детектора $d_k$ ДИНС. $Con(d_k)$ определяет необходимые и достаточные условия возбуждения нейрона-детектора $d_k$., что, в свою очередь, определяет необходимые и достаточные признаки классификации или идентификации распознаваемых образов. Именно эти условия в своей совокупности позволяют преодолеть порог возбуждения (ПВ) нейрона-детектора. Таким образом, $Con(d_k)$ можно рассматривать как фильтр входящей информации, настраиваемый на определенные признаки распознавания в процессе обучения. Следовательно, концепт является информационной моделью дендритного дерева нейрона-детектора.

Тогда процесс обучения нейрона-детектора класса распознавания будет сводиться к: 1) формированию (определению) концепта и 2) минимизации концепта, т.е. нахождению минимального связного вектора входных сигналов. Минимизированный концепт представляет собой аттрактор - устойчивую, т.е. не модифицируемую в процессе дальнейшего обучения, необходимую и достаточную структуру связных элементов входного вектора сигналов, инвариантную аффинным и деформационным преобразованиям структуры распознаваемого контурного изображения. Именно такой концепт будет определять класс распознавания. В этом случае, весовые синаптические коэффициенты приобретают совершенно иной смысл, отличный от традиционного толкования коннекционистской парадигмы.

*Определение 7.* Весовой коэффициент $w_{ik}$ синаптической связи i-го пресинаптического нейрона (элемента входного вектора сигналов $x_i$) с k-м постсинаптическим нейроном будем рассматривать как коэффициент принадлежности данного элемента входного вектора сигналов концепту $Con(d_k)$ нейрона-детектора $d_k$.

Коэффициент принадлежности $w_{ik}$ вычисляется на основе значения функции принадлежности f. Пусть:

$$w_{ik} = f(x_i, Con(d_k)). \quad (8)$$

Для рассматриваемой ДИНС достаточно, чтобы данная функция f была бинарной и принимала значение "0" или "1". В этом случае:

$$w_{ik} = 1, \text{ если } x_i \in Con(d_k)\ ; \quad (9)$$

$$w_{ik} = 0, \text{ если } x_i \notin Con(d_k). \quad (10)$$

При этом:

$$x_i \in Con(d_k), \text{ если: } x_i = mod(x_i) \in Con(d_k);$$

$$\exists!\ E(mod(x_i), <mod(x_j),\ldots, mod(x_k)>) \in Con(d_k), \quad (11)$$

где E – множество отношений пространственной связности моды $mod(x_i) \in Con(d_k)$ и соответствующих мод $<mod(x_j),\ldots, mod(x_k)> \in Con(d_k)$.

$$E=(e_{ij},\ldots, e_{ik}), \quad (12)$$

где $e_{ij}$ – отношение пространственной связности мод $mod(x_i)$ и $mod(x_j)$.

Данное отношение $e_{ij}$ существует для элементов входного вектора сигналов $x_{iI}$ и $x_{jI}$, если:

$$(I(x_i)=I(x_j) \text{ or } I(x_i)\cap I(x_j)) \rightarrow \exists\, e_{ij}, \quad (13)$$

где I – соответствующее значение пространственного индекса.

Следовательно, исходя из выражений (11), коэффициенты принадлежности $w_{ik}$ должны формироваться отдельно для адресной составляющей $a_i$ моды и для ее пространственного индекса $I_i$.

Теперь определим, что значение функции f – принадлежности сигнала $x_i$, поступающего на i-й вход детектора $d_i$, его концепту $Con(d_i)$ (выражение (8)) имеет не просто бинарное значение, как в выражениях (9) и (10), а определяется на основе некоторого статистического значения $q_i$, которое моделирует правило обучения Хебба и позволяет избежать постоянного и резкого переучивания нейрона-детектора (отдельных компонентов его концепта) в каждый момент времени обучения. Для обеспечения сходимости $Con(d_i)$ к аттрактору в процессе обучения введем нелинейную функцию $q_i$ :

$$q_i = (l_i/k)^c, \quad (14)$$

где $q_i$ – значение функции принадлежности f (выражение (8)); $l_i$ – суммарное число поданных сигналов $x_i$ на i-й вход детектора $d_i$ в моменты времени обучения tr. В данные моменты времени $x_i$ поступает синхронно с сигналом обучения z; k – общее количество циклов подачи векторов входных сигналов $\overline{X}$ на входы $d_i$ в моменты времени обучения tr; c - константа (0 < c < 1), определяющая нелинейность $q_i$. Эта нелинейность замедляет процесс сходимости концепта к аттрактору, что позволяет устранить постоянное и резкое переучивание нейрона.

Для определения принадлежности $x_i$ концепту $Con(d_i)$ введем статистический порог принадлежности σ < 1 (рис. 4).

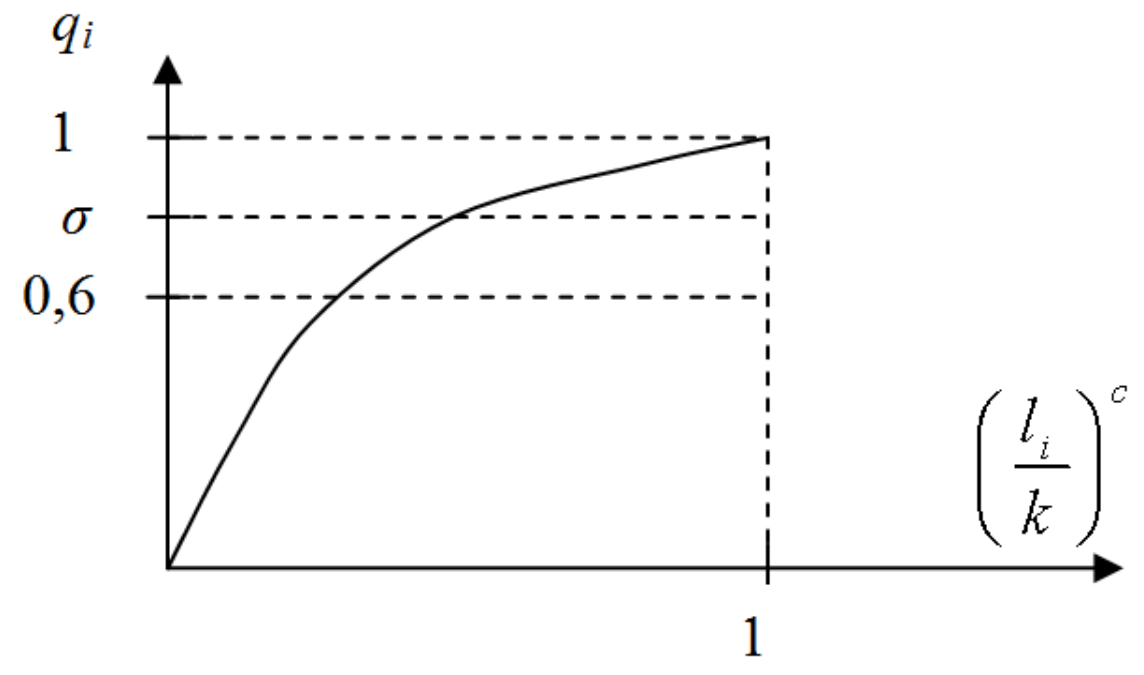

Рис. 4. Функция принадлежности $q_i = (l_i/k)^c$

Тогда, первое правило выражения (11) будет определяться следующим соотношением:

$$q_i< \sigma, \text{ то } w_i =0 \text{ и } \rightarrow x_i(a) \in Con(d_i);$$
$$q\geq \sigma, \text{ то } w_i =1 \text{ и } \rightarrow x_i(a) \in Con(d_i). \quad (15)$$

В данном случае возникает проблема выбора значения порога принадлежности σ и связанная с ней проблема доверия к данному выбору. Чем ближе значение σ будет к 1, тем с большей уверенностью можно утверждать, что данный входной сигнал принадлежит $Con(d_i)$, тем менее "пластичным" будет $Con(d_i)$ в процессе обучения и наоборот. Данную проблему можно назвать проблемой надежности-пластичности $Con(d_i)$.

Такими образом, модификация $Con(d_i)$ в процессе обучения связана со следующими изменениями:

1) Изменением (уменьшением либо увеличением) количества либо значения характеристических либо структурных мод в $Con(d_i)$. Эти изменения позволяют сформировать необходимые и достаточные инвариантные, относительно аффинных и деформационных преобразований контура изображения, признаки распознавания и, в конечном итоге, сформировать ДИНС, представляющую собой классификационную структуру нейронов-детекторов.

2) Изменением в процессе обучения пространственной связности мод $Con(d_i)$. Эти изменения связаны с изменением значений соответствующих пространственных индексов. В конечном итоге данный процесс может привести к проверке только самого факта пространственной связности без привязки к значениям пространственных индексов: конкретному местоположению распознаваемого изображения или элемента изображения в поле восприятия. Это позволяет сформировать концепт инвариантный к местоположению его структурных элементов.

**2. Процедура встречного обучения.** Рассмотрим более детально процедуру обучения отдельных нейронов-детекторов на вторичном уровне обработки информации в ДИНС, связанную с модификацией концепта.

Концепция взаимного или встречного обучения нейронов-детекторов ДИНС является обобщением и расширением известных парадигм обучения "с учителем", "без учителя", с подкреплением.

Введем следующие определения [28].

*Определение 8.* Систему ДИНС, осуществляющую формирование презентаций (от лат. praesentatio – представление) воспринимаемого образа, будем называть презентативной системой PrS.

В данном контексте презентация – это не способ предъявления информации кому-либо, а модель внутреннего представления в ДИНС воспринимаемого образа. Построение презентаций в рассматриваемой ДИНС осуществляется на первичном и вторичном уровнях обработки информации.

*Определение 9.* Систему ДИНС, осуществляющую формирование вторичных презентации - репрезентаций воспринимаемых образов в другом формальном алфавите, будем называть репрезентативной системой RpS [29].

Именно репрезентативная система RpS позволяет определить результат распознавания модели воспринимаемого образа (презентации) и осуществить процесс встречного обучения нейронов-детекторов презентативной системы PrS. Таким образом, сопоставление конкретной презентации буквы алфавита репрезентативной системы - адреса нейрона-детектора RpS и будет являться и итогом распознавания и моделью обучения.

На рис. 5 представлены: репрезентативная система с выделенным единичным нейроном, участвующим в процессе обучения; множество презентативных подсистем с двумя выделенными нейронами в двух различных подсистемах, участвующих в обучении; перцептивная система (система восприятия) с двумя выделенными подсистемами, одновременно воспринимающими своими рецептивными полями (выделены овалом) соответствующие вектора входных сигналов $X_1$ и $X_2$. Презентативные подсистемы, связанные со своими перцептивными подсистемами, формируют презентации – модели воспринимаемых образов, представленные ансамблем возбужденных нейронов-детекторов (выделены овалами в презентативных подсистемах).

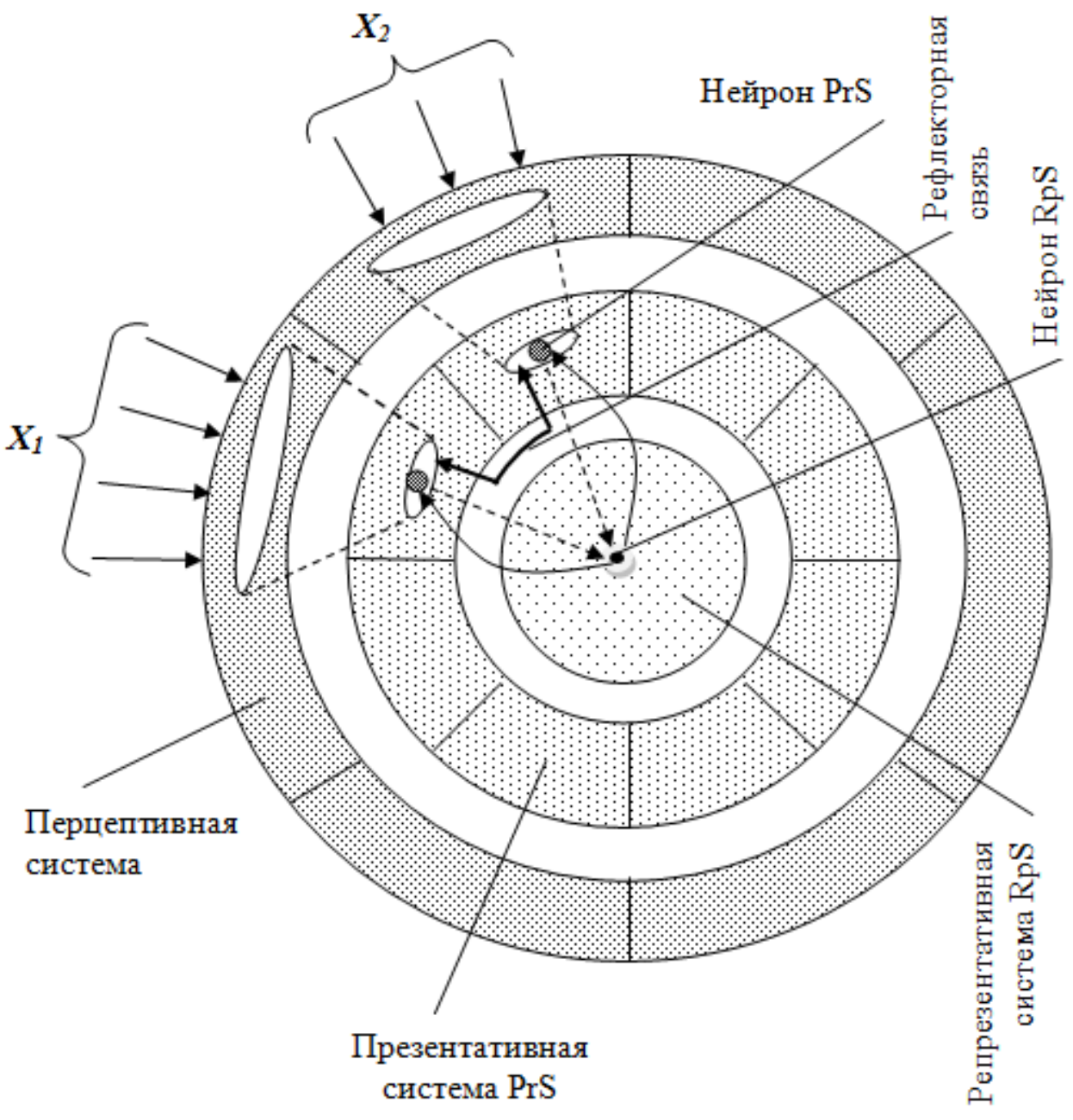

Рис. 5. Схема встречного обучения нейронов презентативных подсистем и репрезентативной подсистемы

Синхронное возбуждение выделенных нейронов в презентативных подсистемах определяет формирование сигналов возбуждения (обозначены пунктирными направленными связями) и процесс обучения выделенного нейрона в репрезентативной подсистеме, который, в свою очередь, формирует встречные управляющие сигналы обучения z (обозначены сплошными направленными связями) для обучения соответствующих нейронов презентативных подсистем. Обученные нейроны формируют ансамбль взаимосвязанных нейронов в репрезентативной системе и презентативных подсистемах. Возбуждение обученных нейронов в любой системе ведет к возбуждению всего ансамбля обученных нейронов. Наиболее важные (или наиболее часто используемые) связи между нейронами различных презентативных подсистем могут становиться короткими (рефлекторными) без участия нейронов репрезентативной системы.

Таким образом, управляющий сигнал обучения z нейрона-детектора $d_i$ формируется нейронами-детекторами других систем, например, репрезентативной системы RpS [29]. Рассмотрим упрощенную пошаговую процедуру обучения нейрона-детектора $d_i$ презентативной системы PrS под управлением нейрона-детектора $d_i^*$ репрезентативной системы RpS. Назовем данные нейроны-детекторы по типу процедуры обучения – встречными.

В процессе встречного обучения детектор PrS должен установить взаимосвязь с соответствующим детектором RpS. Схема установления данной взаимосвязи представлена на рис. 6.

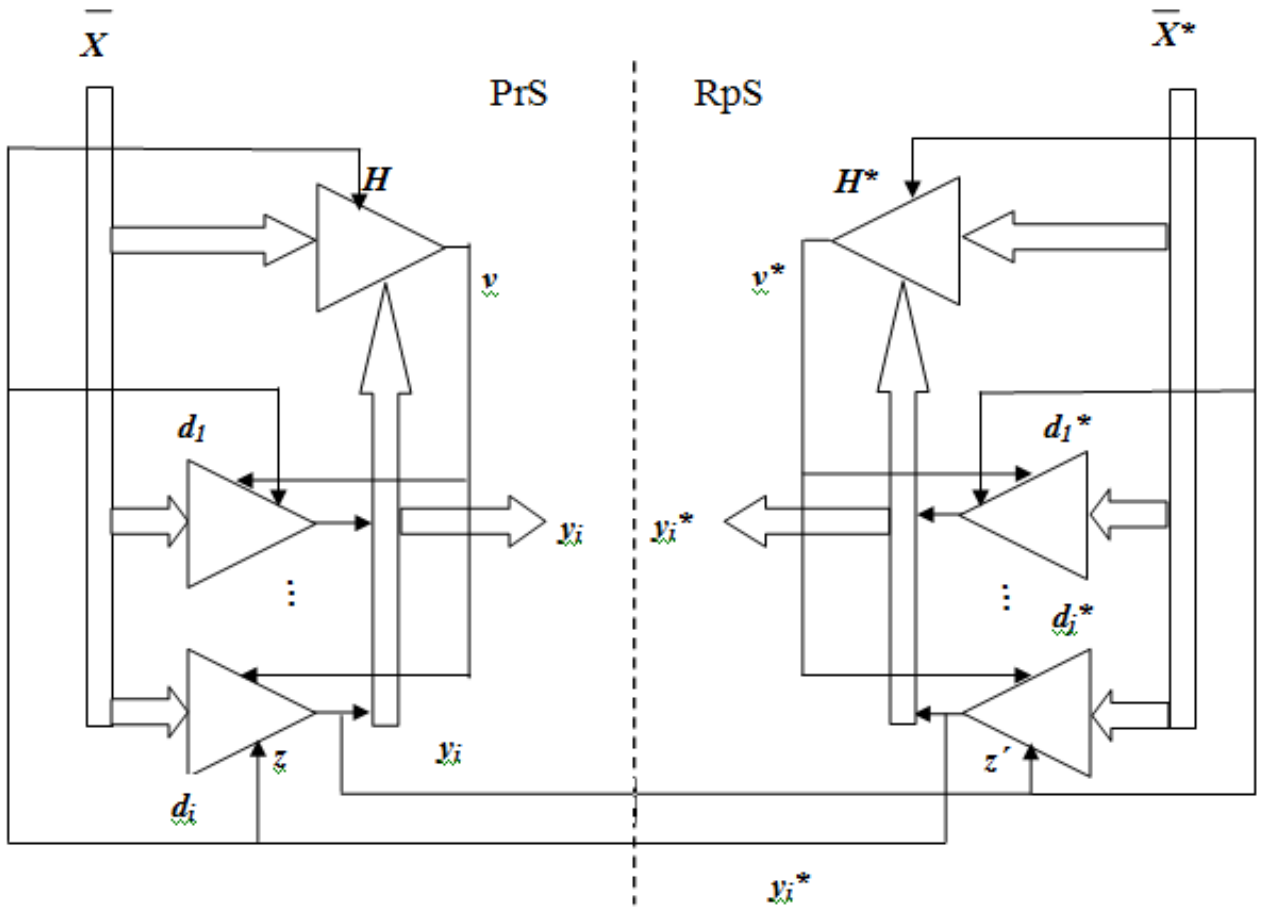

Рис. 6. Схема установления взаимосвязи детекторов $d_i$ и $d_i^*$ в процессе встречного обучения

1. Предположим, что в начальный момент времени детектор $d_i$ не активирован и находится в состоянии "свободен". Данное состояние характеризуется тем, что при поступлении на его входы любого вектора сигналов $\overline{X}$ не происходит его возбуждения т.к. Con($d_i$) не сформирован. Также детектор не распознает сигналов обучения z ввиду того, что $d_i$ еще не связан с $d_i^*$ - обучающим детектором RpS. Данная связь формируется в результате запоминания детектором $d_i$ "адреса" $d_i^*$.

Для активации детектора $d_i$ необходимо осуществить его "захват". Для этого нужен командный нейрон H, управляющий процессом активизации всех нейронов модуля. На входы H поступают все выходные сигнала нейронов модуля, сигналы обучения z и входной вектор сигналов $\overline{X}$. Если в модуль поступает $\overline{X}$ и сигнал z, но ни один из нейронов модуля не возбуждается, то на выходе H формируется сигнал v

"захвата" свободного нейрона-детектора. Следовательно, командный нейрон Н является нейроном новизны. Допустим, что ближайший свободный к Н детектор $d_i$, получив управляющий сигнал v и сигнал обучения z – выходной сигнал $y^*$ нейрона $d_i^*$, воспринимает входной вектор $\overline{X}$ как вектор распознавания и запоминает его в качестве исходного $Con(d_i)$, а сигнал $y^*$ запоминает в качестве связанного сигнала обучения z. Таким образом, $d_i$ становится детектором – идентификатором примера $\overline{X}$. В случае, если дальнейшего обучения детектора $d_i$ не произойдет, он будет возбуждаться исключительно при поступлении на его входы вектора $\overline{X}$.

2. Детектор $d_i$ возбужден, но еще не установлена его обратная связь с обучающим детектором $d_i^*$. Детектор $d_i$ посылает сигнал $y_i$ в модуль RpS и одновременно возбужденный с $d_i$ детектор $d_i^*$ в RpS воспринимает данный сигнал как встречный управляющий сигнал обучения z′. В этом случае значение сигнала $y_i$ запоминается детектором $d_i^*$ и таким образом устанавливается взаимосвязь $d_i$ и $d_i^*$. Таким образом, детекторы $d_i$ и $d_i^*$ обучают друг друга.

3. При возникновении на входе $d_i$ вектора $\overline{X} \cap Con(d_i)$ и наличии управляющего сигнала обучения z, происходит обучение $d_i$ в результате коррекции $Con(d_i)$.

4. В случае если при возникновении z на входы $d_i$ поступает вектор $\overline{X}$, который не пересекается с $Con(d_i)$, то детектор $d_i$ не возбуждается. В этом случае происходит "захват" и возбуждение нового детектора с альтернативным для данного сигнала z концептом.

5. Если на входы $d_i$ поступает вектор $\overline{X} \cap Con(d_i)$, но одновременно возникающий сигнал обучения z не совпадает с запомненным значением $y_i^*$, то детектор $d_i$ не возбуждается (тормозится). Тогда происходит "захват" и возбуждение нового детектора $d_k$ с $Con(d_k)=\overline{X}$ и новым значением сигнала z.

По такому же алгоритму работает и обратная процедура, когда ведущий сигнал обучения поступает от детектора PrS. В этом случае первоначальный "захват" детектора $d_i$ управляющим сигналом v осуществляется без сигнала z, т.е. происходит самообучение детектора $d_i$.

**3. Составляющие реакции нейрона-детектора.** Представляемая информационная модель описывает процессы преобразования информации на вторичном уровне в нейроне-детекторе ДИНС, которые гипотетически могут быть реализованы известными физиологическими механизмами в биологическом нейроне.

В соответствии с детекторным подходом, для построения информационной модели нейрона-детектора ДИНС определим основные информационные элементы, составляющие его реакцию $y_i$:

$$y_i(a,b,c,I), \qquad (16)$$

где $y_i$ – комплексная реакция нейрона-детектора; a – адресная составляющая реакции; I – пространственный индекс реакции; b – уровень возбуждения нейрона-детектора; c – тип возбуждения нейрона-детектора.

Формирование адресной составляющей и пространственного индекса было рассмотрено выше.

Уровень возбуждения нейрона-детектора b моделирует уровень возбуждения: частотную либо амплитудную (градуальную) реакцию биологического нейрона. Значение модельной составляющей реакции b равно линейной сумме числа всех мод концепта Con и в точности соответствует порогу возбуждения $g_i$ нейрона-детектора $d_i$ (выражение (17)). Это допущение хоть и существенно упрощает реальные процессы нелинейной суммации ВПСП в биологическом нейроне, но не исключает вариативность уровня возбуждения b в модельном нейроне. Однако для целей моделирования эти упрощения оправданы.

$$b = (\sum_{j=1}^{k} mod(x_j) \in Con(d_i)) = g_i, \qquad (17)$$

где $mod(x_j)$ - любая мода, принадлежащая концепту; $g_i$ - порог возбуждения данного нейрона-детектора.

Тогда:

$$(\sum_{j=1}^{k} mod(x_j) \in Con(d_i)) = g_i) \to \exists\, y_i(b). \qquad (18)$$

Из данного выражения также видно, что процесс обучения нейрона ведет к изменению пороговой величины $g_i$, что также является биологически обоснованным [4].

Рассматриваемая составляющая b необходима для реализации процесса WTA - конкуренции одновременно возбужденных нейронов-детекторов в одном модуле (одной карте детекторов) ДИНС для определения одного лидера в модуле, сигналы от которого будут участвовать в дальнейшей обработке информации. Кроме того, данная составляющая необходима и для организации структурирования карт детекторов в процессе самообучения, который будет рассмотрен ниже. Назовем процесс WTA-конкуренции между нейронами-детекторами ДИНС - α-конкуренцией [30].

Действительно, у нейронов-детекторов одного модуля не может существовать двух одинаковых концептов, но может быть случай, когда, например, $Con(d_1) \subset Con(d_2)$. В этом случае будут одновременно возбуждены детекторы $d_1$ и $d_2$, но значение $y_1(b)$ детектора $d_1$ будет меньше $y_2(b)$ детектора $d_2$.

При α-конкуренции выходные сигналы $y_i(b)$ одновременно возбужденных нейронов-детекторов сравниваются в компараторах С каждого из детекторов, где, в случае превышения какого либо из внешних сигналов $y_i(b)$ над своим собственными, генерируется внутренний управляющий сигнал торможения h, который "сбрасывает" возбуждение детектора переводя его в неактивное состояние. На рис. 7 показана схема формирования сигнала h в детекторе $d_1$ при условии $y_2(b) > y_1(b)$.

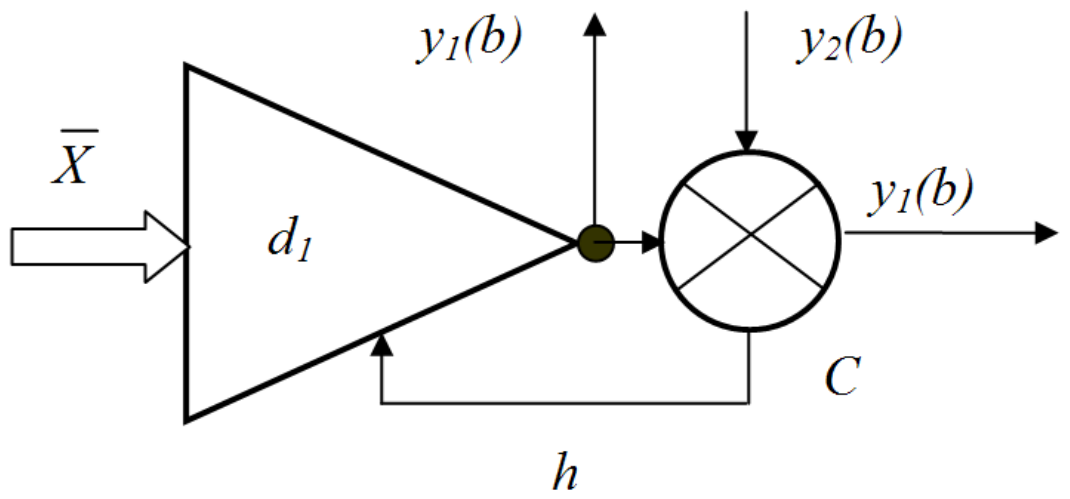

Рис. 7. Схема формирования сигналов h и $y_1(b)$

В случае если $y_1(b) > y_2(b)$, то тормозится детектор $d_2$. Тогда компаратор C детектора $d_1$ передает сигнал $y_1(b)$ на выход нейрона-детектора. Таким образом осуществляется процесс латерального торможения для реализации α-конкуренции нейронов-детекторов в модуле.

Для биологических нейронов, возможно, что передача сигналов возбуждения для реализации процесса α-конкуренции между нейронами в нейронном модуле по горизонтальным связям может осуществляться с использованием аксо-аксональных синапсов.

При ассоциативном распознавании, которое будет рассмотрено ниже, α-конкуренция происходит только между ассоциативно возбужденными нейронами-детекторами.

Уровень возбуждения b нейрона-детектора ДИНС и тип его возбуждения c в совокупности моделируют частотную реакцию биологического нейрона. Известно, что биологические нейроны могут демонстрировать несколько различных видов активности [31]. Различные виды реакций биологических нейронов-детекторов, например: пачечная (известная также как "берстовая" или "залповая") реакция, регулярная, нерегулярная или периодическая, спонтанная активность и др., вероятно связаны с их внутренним состоянием и отражают динамические свойства нейронов [5].

Однако предположим, что различные виды реакции нейрона-детектора, с информационной точки зрения, зависят от типа его возбуждения, который, в свою очередь, определяется условиями возбуждения. Этими условиями могут быть: вид реакции пресинаптических нейронов; полнота вектора входных сигналов, определяющего необходимые и достаточные условия возбуждения нейрона-детектора; наличие входных управляющих сигналов z, поступающих, например, от встречных нейронов других систем.

Для нейронов-детекторов ДИНС определим два типа возбуждения, под которыми будем понимать:

1) прямое возбуждение – возбуждение нейронов-детекторов любого типа при наличии всех необходимых и достаточных элементов входного вектора сигналов, определяемых концептом. В этом случае возбуждение нейрона-детектора осуществляется с учетом значения порога возбуждения $g_i$ в соответствии с выражением (17). Прямое возбуждение нейрона-детектора ДИНС моделирует пачечную активность биологического нейрона;

2) ассоциативное возбуждение возможно только нейронов-детекторов производных структур. В этом случае значение порога возбуждения $g_i$ не учитывается. Предположим, что ассоциативное возбуждение нейронов-детекторов ДИНС моделирует регулярную активность биологических нейронов и может возникать в следующих случаях:

**А.** Если все входные сигналы, необходимые и достаточные для возбуждения, т.е. принадлежащие Con, являются реакциями прямого возбуждения пресинаптических нейронов-детекторов непроизводных структурных элементов, но при этом часть из них отсутствует, т.е. входной вектор – неполный. В этом случае Con выступает в роли эталонной структуры ансамбля пресинаптических нейронов необходимых для ассоциативного возбуждения данного нейрона-детектора. Таким образом, если во входном векторе присутствуют хотя бы два сигнала, активизирующих связные структурные моды принадлежащие Con, что необходимо в соответствии с выражением (11), то нейрон-детектор ассоциативно возбуждается (в его реакции будет присутствовать значение ассоциативного типа возбуждения) и его уровень возбуждения b будет равен сумме числа элементов входного вектора сигналов, принадлежащих Con, вне зависимости от значения порога возбуждения $g_i$.

**Б.** Если на входе нейрона-детектора отсутствуют все входные сигналы, но есть управляющий сигнал возбуждения z от встречного нейрона-детектора, то данный нейрон ассоциативно возбуждается с уровнем возбуждения b равным сумме числа всех элементов Con.

**В.** Для нейронов-детекторов сложных производных структур, где входные сигналы могут поступать от нейронов-детекторов более простых производных структур, также выполняются условия ассоциативного возбуждения **А** или **Б**. Однако, даже если входной вектор сигналов полный (соответствует Con), но хотя бы один входной сигнал является реакцией ассоциативного возбуждения, то данный нейрон также возбуждается ассоциативно.

Ассоциативное возбуждение нейронов-детекторов определяет ассоциативное распознавание образов, что с психологической точки зрения, по сравнению с прямым возбуждением, означает более низкую степень уверенности в результате распознавания - гипотезу распознавания.

Таким образом, составляющая реакции c нейрона-детектора определяет тип его возбуждения и равна, например, "1" при прямом возбуждении либо "0" при ассоциативном возбуждении.

**4. Информационная модель нейрона-детектора ДИНС.** Информационная модель нейрона-детектора ДИНС состоит из модели дендритного дерева, модели обучения и модели формирования реакции.

**4.1. Информационная модель дендритного дерева.** Структурная схема информационной модели дендритного дерева нейрона-детектора ДИНС для процесса распознавания входного вектора сигналов представлена на рис. 8.

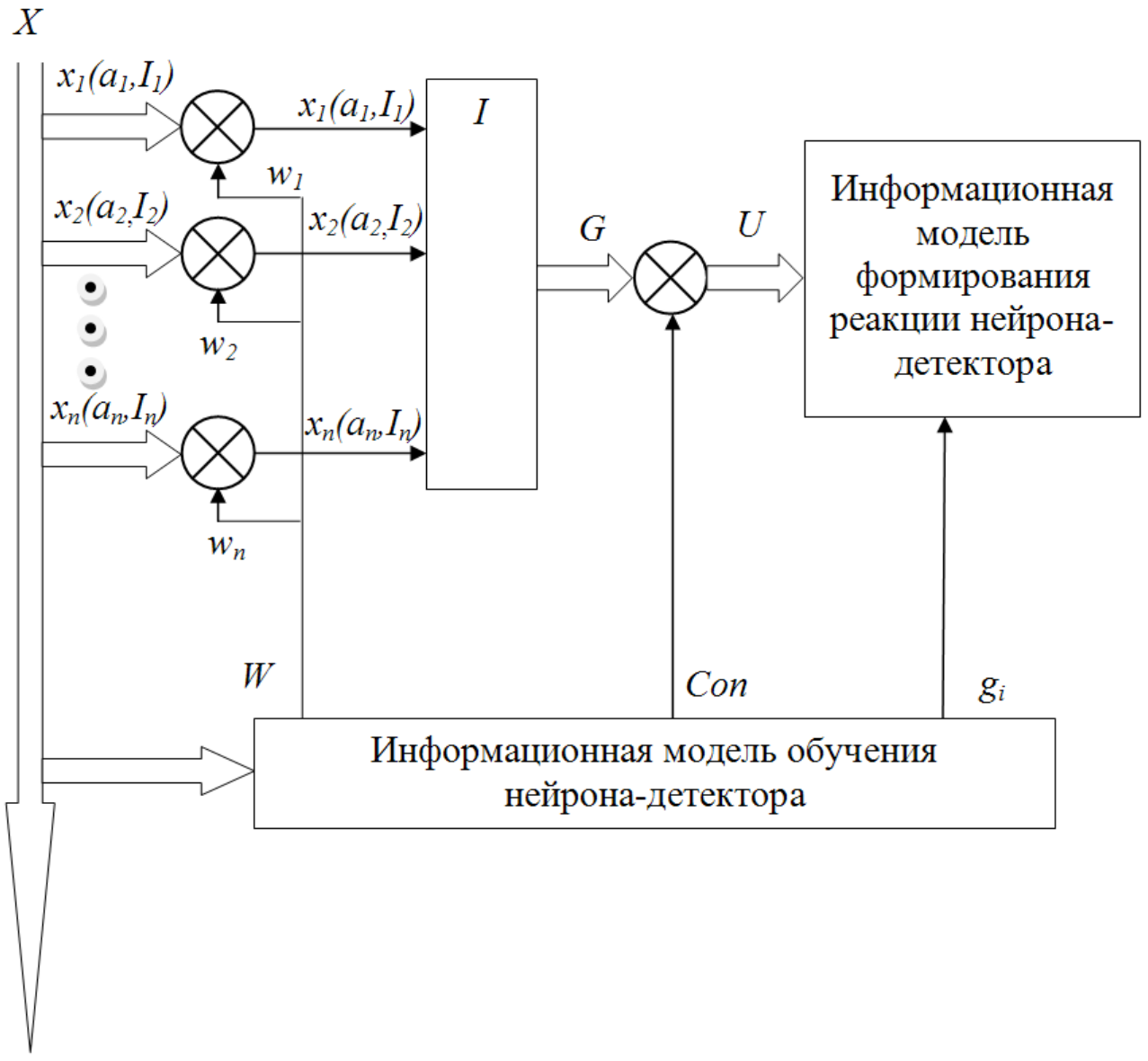

Рис. 8. Структурная схема информационной модели дендритного дерева нейрона-детектора ДИНС

На данной схеме представлены следующие элементы и блоки:

X – входной вектор сигналов;

$x_i(a_i,I_i)$ – i-й элемент входного вектора сигналов с адресной составляющей $a_i$ и пространственным индексом $I_i$;

$w_i=(w_{ia},w_{iI})$ – пара весовых коэффициентов принадлежности двух составляющих: $w_{ia}$ - адресной составляющей $a_i$ и $w_{iI}$ - значения пространственного индекса $I_i$ элемента входного вектора сигналов $x_i$ концепту Con, т.е. данные весовые коэффициенты, в соответствии с принципом абсолютной адресации в ДИНС, связаны не с конкретным синаптическим входом, а со значениями отдельных составляющих моды;

W – множество значений матрицы весовых коэффициентов принадлежности, размерность которой соответствует размерности Con (W формируется в процессе обучения);

I – блок определения пространственной связности элементов входного вектора сигналов (мод), принадлежащих Con;

G – пространственно-связный граф мод;

Con – концепт нейрона-детектора, формируемый в процессе обучения;

U – множество возбуждающих постсинаптических потенциалов

$$U=(u_i(c_i), \ldots, u_k(c_k)), \qquad (19)$$

где $u_i$ – значение возбуждающего постсинаптического потенциала, соответствующего i-й моде, принадлежащей Con, т.е. если выполняются условия, определенные в выражениях: (11), (13), (15), то в этом случае $u_i=1$, в противном случае $u_i=0$;

$c_i$ – значение типа возбуждения пресинаптического нейрона, реакция которого соответствует моде Con, т.е. если пресинаптический нейрон имеет прямое возбуждение, то $c_i=1$, а если ассоциативное – то $c_i=0$;

$g_i$ – значение порога возбуждения нейрона-детектора, формируемое в процессе обучения;

На схеме перечеркнутым кругом обозначаются блоки сравнения – компараторы.

Данная модель в режиме распознавания входного вектора сигналов функционирует следующим образом:

1. При поступлении на входы нейрона-детектора вектора входных сигналов X:

$$X= (x_1(a_1,I_1,c_1), x_2(a_2,I_2,c_2),\ldots, x_n(a_n,I_n,c_n)) \qquad (20)$$

в компараторах каждого входа, на основе соответствующих весовых коэффициентов $w_i$, в соответствии с выражением (15), определяется принадлежность каждой адресной составляющей $a_i$ и пространственного индекса $I_i$ сигнала $x_i$ моде концепта Con, т.е. проверяется выполнение первого правила выражения (11).

2. В случае если $x_i(a_i, I_i) \in$ Con, в блоке I определения пространственной связности элементов входного вектора сигналов на основе значений сигналов $x_i(I_i)$ и выражений (11), (13), (15) осуществляется построение пространственно-связного графа мод G.

3. В последующем компараторе осуществляется проверка выполнения второго правила выражения (11) в результате сравнение графа G с Con. В случае если графы G и Con изоморфны, идентичны или G ∩ Con (например, при ассоциативном распознавании), то формируется множество возбуждающих постсинаптических потенциалов U с соответствующими значениями типа возбуждения $c_i$ (формула (19)), которые являются выходными значениями данной модели.

**4.2. Информационная модель обучения нейрона-детектора ДИНС.** Данная модель реализует модель встречного обучения. Данное обучение на уровне нейрона-детектора заключается в первичном формировании и последующей модификации в процессе обучения: весовых коэффициентов принадлежности адресных составляющих $w_{ia}$ и пространственных индексов $w_{iI}$ входных сигналов концепту Con, самого концепта Con, а также порога возбуждения $g_i$. Первичное формирование каждого из коэффициентов $w_{ia}$ и $w_{iI}$ осуществляется в соответствии с формулами (14) и (15) при синхронном поступлении вектора входных сигналов X, управляющего сигнала обучения z и сигнал "захвата" v (см. рис.6). В это же время при тех же условиях происходит и первичное формирование концепта Con нейрона-детектора в соответствии с формулами (11), (13).

Изменение (модификация) весовых коэффициентов принадлежности $w_{ia}$ и $w_{iI}$ ведет к модификации концепта Con. Данные изменения осуществляется в соответствии с этими же формулами при синхронном поступлении только вектора входных сигналов X и управляющего сигнала обучения z. В начальный момент времени (при инициализации), все значения $w_{ia}$ и $w_{iI}$ матрицы весовых коэффициентов W имеют значения "1".

Первичное формирование и последующая модификация порога возбуждения $g_i$ связаны с изменениями концепта Con (в момент инициализации g=0) и осуществляются при тех же условиях, которые были рассмотрены для коэффициентов принадлежности и концепта Con, в соответствии с выражением:

$$g_i = \sum_{j=1}^{k} u_j \,, \qquad (21)$$

где $u_j$ – ненулевое значение j-го возбуждающего постсинаптического потенциала (выражение (19)).

На рис. 9 представлена структурная схема информационной модели обучения нейрона-детектора ДИНС.

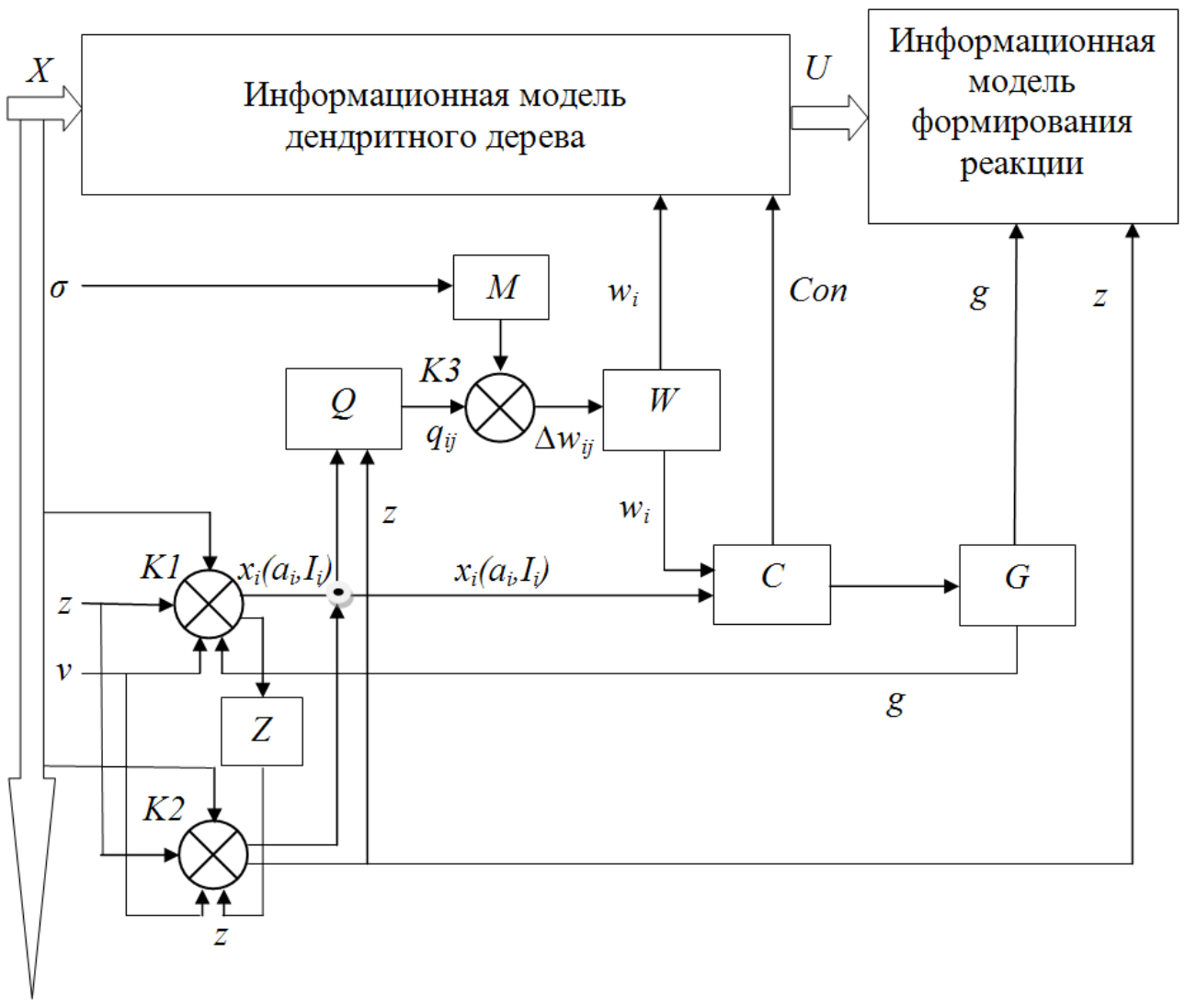

Рис. 9. Структурная схема информационной модели обучения нейрона-детектора ДИНС

На данной схеме изображены следующие информационные блоки и элементы:

X – входной вектор сигналов;

$x_i(a_i,I_i)$ – множество i-х элементов входного вектора сигналов с адресной составляющей $a_i$ и пространственным индексом $I_i$, принадлежащих концепту Con данного нейрона-детектора;

g – значение порога возбуждения нейрона-детектора, формируемое и модифицируемое в процессе обучения, в соответствии с выражением (17), блоком G;

$w_i$ – множество пар ($w_{ia}$, $w_{iI}$) значений весовых коэффициентов принадлежности адресной составляющей $a_i$ и значения пространственного индекса $I_i$ элемента входного вектора сигналов $x_i$ концепту Con,

W – блок формирования, модификации и хранения множества значений матрицы весовых коэффициентов принадлежности – пар ($w_{ia}$, $w_{iI}$);

z – внешний управляющий сигнал обучения, значение которого запоминается в процессе "захвата" нейрона-детектора и хранится в модуле памяти Z. При дальнейшем обучении нейрона-детектора это значение используется для вычисления $q_{ij}$ (выражение (14));

v - внешний управляющий сигнал "захвата" нейрона-детектора;

C – блок формирования, модификации и хранения концепта Con данного нейрона-детектора в соответствии с выражениями (11), (13);

$q_{ij}$ – значение функции принадлежности f (выражение (14)), вычисляемое в процессе обучения и хранимое для каждого элемента пары ($a_i$,$I_i$) в блоке Q;

$\Delta w_{ij}$ – модифицированные значения элементов пар ($w_{ia}$, $w_{iI}$), полученные в результате сравнения в компараторе K3 модифицированных значений $q_{ij}$ и σ (выражение (15));

σ – задаваемое значение ($0{,}6 \leq \sigma < 1$) статистического порога принадлежности $q_i$ адресной составляющей $a_i$ и значения пространственного индекса $I_i$ элемента входного вектора сигналов $x_i$ концепту Con (выражение (15)). Данное значение хранится в блоке памяти M;

K1 – компаратор, определяющий момент "захват" нейрона-детектора при наличии: множества входных сигналов $x_i(a_i,I_i)$; управляющих сигналов z и v; а также при значении порога возбуждения g=0, что свидетельствует об отсутствии концепта Con данного нейрона-детектора;

K2 – компаратор, определяющий каждый момент обучения нейрона-детектора при наличии: множества входных сигналов $x_i(a_i,I_i)$, управляющего сигнала z, совпадающего со значением, запомненным в модуле памяти Z; а также и при отсутствии сигнала v.

Таким образом, память в нейроне-детекторе реализуется следующими компонентами: памятью в блоке Q значений функции принадлежности $q_{ij}$ для каждого элемента пары ($a_i$,$I_i$); памятью в блоке W множества пар значений весовых коэффициентов принадлежности ($w_{ia}$, $w_{iI}$); памятью в блоке C концепта Con; памятью в блоке M значения статистического порога принадлежности σ; памятью в блоке G значение порога возбуждения нейрона-детектора g; памятью в блоке Z значения управляющего сигнала обучения z.

**4.3. Информационная модель формирования реакции нейрона-детектора.** Модель формирования реакции $y_i$(a, b, c, I) нейрона-детектора $d_i$ отражает процессы преобразования информации в "теле" и "аксоне" нейрона ДИНС. Функционирование данной модели проиллюстрировано на структурной схеме (рис. 10), где, в дополнение к элементам и блокам, представленным на рис. 8 и 9, представлены следующие элементы и блоки:

$u_i(c_i{=}1)$ - ненулевое значение i-го возбуждающего постсинаптического потенциала (выражение (19)) от пресинаптического нейрона с прямым типом возбуждения: $c_i$=1;

$u_i(c_i{=}1/0)$ - ненулевое значение i-го возбуждающего постсинаптического потенциала от пресинаптического нейрона с прямым ($c_i$=1) либо ассоциативным ($c_i$=0) типом возбуждения;

$\sum_{j=1}^{k} u_j \sum_{j=1}^{k} u_j$ – блок суммирования возбуждающих постсинаптических потенциалов;

b – уровень возбуждения нейрона-детектора (выражение (17));

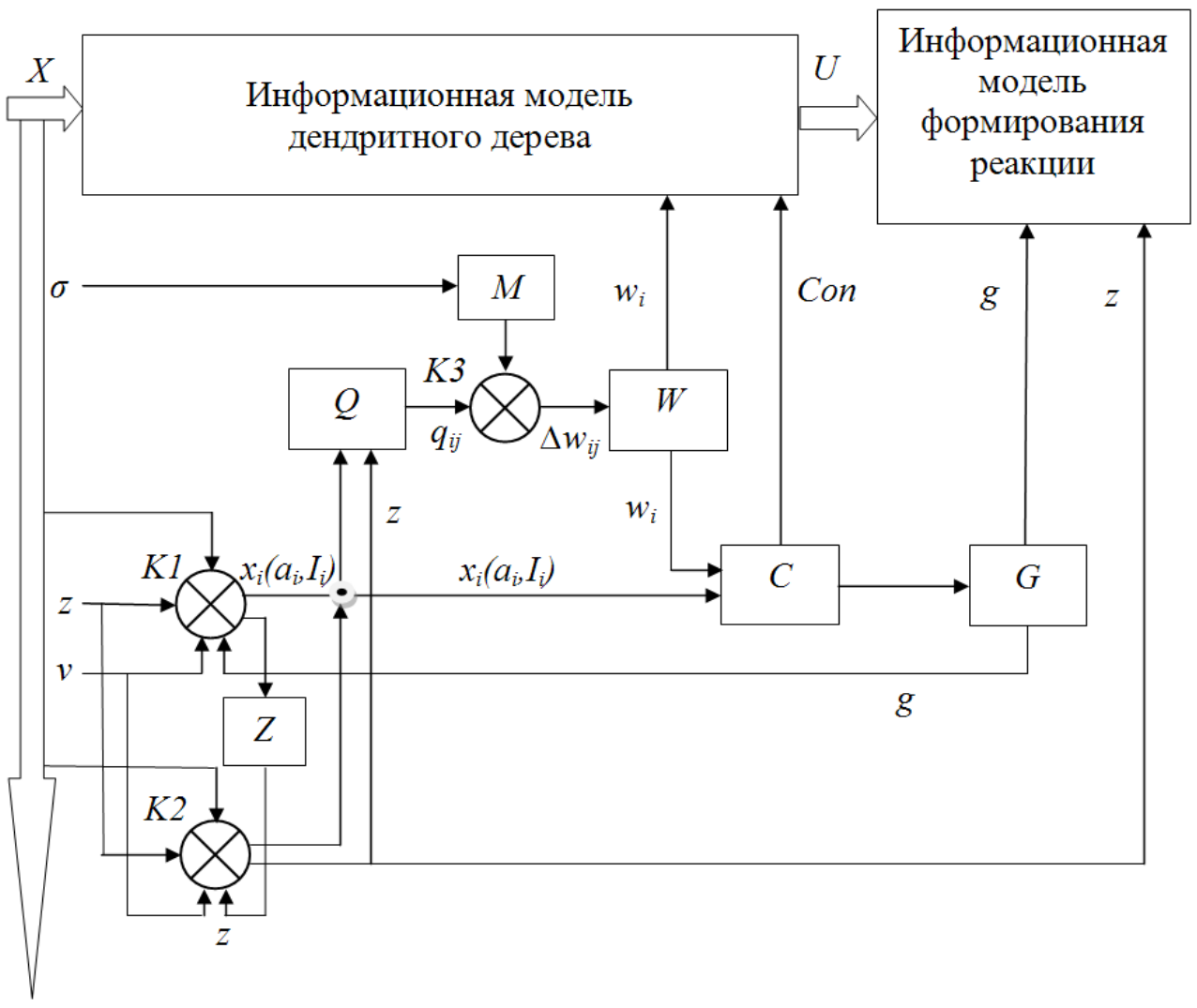

Рис. 10. Структурная схема информационной модели формирования реакции $y_i$(a, b, c, I) нейрона-детектора $d_i$ ДИНС

K1 – компаратор, в котором осуществляется проверка условий прямого возбуждения нейрона: если b=g при условии, что все составляющие $u_i(c_i)$ имеют значение $c_i$=1, то формируется составляющая реакции нейрона-детектора y(b,c=1);

K2 - компаратор, в котором осуществляется проверка условий ассоциативного возбуждения нейрона при котором формируется составляющая реакции нейрона-детектора y(b,c=0), где b=g:

1) если b=g при условии, что все или хотя бы одна составляющая $u_i(c_i)$ имеют значение $c_i$=0, и z=0;

2) если b<g при любых значения $c_i$ и z=0;

3) если b=0, т.е. возбуждающие постсинаптические потенциалы отсутствуют, но z≠0;

y(b,c) – выходной сигнал нейрона-детектора, который подается на управляющие входы соответствующих компараторов всех нейронов данного модуля (карты) для осуществления процесса α-конкуренции (рис.7). Нейроны-победители определяются отдельно для группы ассоциативно возбужденных нейронов и для группы нейронов с прямым возбуждением. Далее, в соответствии с правилом WTA, приоритет получает нейрон с прямым возбуждением (если он существует), а все остальные нейроны тормозятся (ВПСП становятся равными "0") собственными (внутренними) управляющими сигналами h, которые генерируется компараторами K3;

K3 – компаратор, в котором осуществляется процесс приоритетной α-конкуренции на основе сравнение собственного сигнала y(b,c) со входными сигналами y'(b,c), поступившими от других возбужденных нейронов данного модуля;

A – модуль памяти, где хранится адресная составляющая реакции нейрона-детектора y(a). Адреса нейронов-детекторов в модуле задаются в процессе инициализации;

I – блок формирования значения пространственного индекса I (см. рис.1) для его инкапсуляции в структуру реакции нейрона-детектора;

M - блок инкапсуляции (связывания) структурных компонент в комплексную реакцию нейрона-детектора;

y(a,b,c,I) – комплексная реакция нейрона-детектора.

**5. Формирование производных характеристик. Нейроны-анализаторы ДИНС.** Очевидно, что наш мозг способен не только "мозаично" воспринимать отдельные информационные признаки внешних образов и объединять их в реакции детекторов, но и сравнивать эти признаки, осуществлять анализ воспринимаемой информации.

В процессе распознавания образов ДИНС необходимо осуществлять не только обобщение признаков для формирования инвариантных концептов нейронов-детекторов класса распознавания, но и выделение специфических признаков для формирования в процессе обучения концептов нейронов-детекторов подклассов и нейронов-детекторов примеров класса (подкласса) распознавания. Эти специфические признаки являются производными от первичных признаков (непроизводных характеристик) восприятия и их возможно получить только в процессе анализа (сравнения) непроизводных характеристик структурных элементов распознаваемого образа. Информация об этих признаках может быть представлена в виде значений производных характеристических мод в концепте структурного нейрона-детектора. Данные моды формируют дополнительную сильную связность модальных групп структурных элементов и определяют существенные параметры фильтрации входной информации нейрона.

**5.1. Производные характеристики.** Непроизводные характеристики могут иметь как количественные значения, упорядоченные на различных линейных шкалах измерений (например: длины отрезков, местоположение точек изображения либо окон восприятия в Декартовой системе координат для изображений КМ, величины углов и направления ориентации отрезков в полярной системе координат и др.), так и качественные значения (например, бинарные значения типа: "Да-Нет" к которым, в частности, относится характеристика определения замкнутости либо разомкнутости контура изображения). Очевидно, что в процессе обучения при модификации концепта нейрона-детектора, например, класса распознавания, многие из данных характеристических мод будут удалены из концепта. Это может привести к потере уникальности концепта и возникновению конфликтных ситуаций при распознавании образов, когда, например, α-конкуренция не приведет к определению нейрона-победителя.

Для того чтобы уникальность концепта была сохранена, во входном векторе сигналов должны быть другие элементы, также имеющие качественные либо количественные значения, которые будут более устой-чивы либо инвариантны к изменению количественных характеристик конкретных структурных элементов, и, в то же время, будут обеспечивать уникальность концепта распознавания нейрона-детектора. Этими элементами могут быть реакции нейронов-детекторов производных или вторичных характеристик структурных элементов. Предположим, что к производным характеристикам объектов КМ, например, относятся: качественные характеристики отношений количественных непроизводных характеристик (например, бинарные отношения: "больше – меньше", "по часовой стрелке – против часовой стрелки", "вправо – влево", и др.); количественные характеристики отношений количественных непроизводных характеристик (например, длин отрезков, величин углов, размеров изображения и др.). Например, качественная производная характеристика, которая может обозначать выпуклость либо вогнутость контура изображения, формируется в результате сравнения направлений ориентации отрезков – векторов при детектировании углов с учетом выбранной точки начала и направления последовательного обхода контура изображения. В этом случае наличие определенных значений данной характеристики – связных мод в структуре концепта определяют неизменность либо изменение направления ориентации последовательно расположенных и пространственно связанных векторов в контуре изображения.

Таким образом, типы производных характеристик должны относиться не к отдельному структурному элементу контура изображения, а ко всему изображению в целом либо определенной подструктуре изображения. Процесс формирования производных характеристик обладает существенными отличиями от процесса формирования непроизводных характеристик:

1) формирование значений данных характеристик происходит последовательно в результате выполнения бинарных операций сравнения значений непроизводных характеристик, в общем случае устанавливающих между ними математические отношения равенства или порядка;

2) формирование производных характеристик вариативно и ситуационно, т.е. имеет структурно-временную зависимость. Это значит, что одни и те же нейроны, осуществляющие сравнение значений непроизводных характеристик, в разные моменты времени могут обрабатывать однотипную информацию, относящуюся к разным элементам одного изображения либо к разным изображениям.

**5.2. Нейроны-анализаторы и ассоциативные моды.** Для формирования значений производных характеристик должны существовать нейроны нового типа, осуществляющие операцию сравнения реакций (а именно - адресных составляющих реакций) нейронов-детекторов непроизводных характеристик. Этими нейронами являются нейроны-анализаторы.

*Определение 10.* Нейронами-анализаторами ДИНС называются нейроны, осуществляющие попарное сравнение упорядоченных на определенной шкале значений адресных составляющих реакций пресинаптических нейронов-детекторов непроизводных характеристик структурных элементов распознаваемого образа либо всего образа в целом.

Очевидно, что для сравнения адресных составляющих реакций нейронов-детекторов непроизводных характеристик, адреса этих нейронов и, следовательно, сами нейроны должны быть линейно упорядочены в соответствующих модулях (картах), образуя количественные (интервальные, метрические) либо качественные (порядковые, неметрические, дихотомические и недихотомические) шкалы значений данных характеристик. Тогда модуль разности двух адресов нейронов-детекторов, упорядоченных на количественной шкале, может определять значение производной количественной характеристики. Подобная упорядоченность нейронов рассматривалась, например, в работах Е.Н. Соколова и С. Зеки в процессе исследования системы восприятия цвета [32].

Последовательный процесс формирования производных характеристик моделирует нейропсихологический процесс активного восприятия распознаваемых образов - активный перцептивный акт, изучению которого посвящено множество научных исследований А.Л. Ярбуса, А.Р. Лурия, Е.Д. Хомской и др. Существенную роль в данном процессе играет система внимания, которая фиксирует фокус внимания и зрительный фокус на структурных элементах изображения - критических точках, которые являются наиболее информативными и потому критически важными для распознавания. Непроизводные характеристики именно этих критических точек и будут сравниваться нейронами-анализаторами [33].

Предположим, что выбор двух структурных критических точек для осуществления сравнения значений их характеристических мод и получения значения производной характеристики осуществляется системой внимания случайным образом. Следовательно, значения пространственных индексов I мод непроизводных характеристик уже не может использоваться в качестве базы связности для этих мод. Однако так как выбор любых двух критических точек осуществляется последовательно, то в качестве базы связности могут использоваться новые информационные элементы в структуре реакции нейрона - последовательные метки фокуса внимания $M_k(t_{ex})$, которые формируются системой внимания в процессе активного перцептивного акта и размещаются (инкапсулируются) в структуре пространственного индекса I в течение времени экспозиции образа $t_{ex}$. В этом случае пространственный индекс $I(h_{ij})$ i-й непроизводной характеристической моды $h_{ij}$, принадлежащей модальной группе j-й структурной моды $x_j$ концепта вторичного (производного) структурного нейрона-детектора, будет иметь следующую структуру:

$$I(h_{ij}) = ( W^{i}_{j*k} (x_j), M_k(t_{ex})), \qquad (22)$$

где индекс пространственного окна, в котором нахо-дится данный структурный элемент – $W_{j*k}^{i}(x_j)$, а $M_k(t_{ex})$ – метка фокуса внимания с индексом последовательности k во временном интервале экспозиции $t_{ex}$.

Тогда две последовательные метки $I(M_k(t_{ex}))$ и $I(M_{k+1}(t_{ex}))$ будут определять связность соответствующих мод непроизводных характеристик, и эта связность будет реализована в виде производной характеристической моды mod( $h'_{jl}$ ):

$$\left[\mathrm{mod}(h_{ij})\left\langle I\left(M_k\left(t_{ex}\right)\right)\right\rangle, \mathrm{mod}(h_{i\ell})\left\langle I\left(M_{k+1}\left(t_{ex}\right)\right)\right\rangle\right] \to \to \exists e_{i\ell} \to \mathrm{mod}(h'_{i\ell})\left\langle I\left(M_k\left(t_{ex}\right), M_{k+1}\left(t_{ex}\right)\right)\right\rangle, \quad (23)$$

где mod ($h_{ij}$) – i-ая характеристическая мода, принадлежащая модальной группе j-й структурной моды; mod ($h_{il}$) - i-ая характеристическая мода, принадлежащая модальной группе l-й структурной моды (равенство индексов i определяет, что данные характеристические моды относятся к одному и тому же типу характеристики); $I\left(M_k\left(t_{ex}\right)\right)$ - пространственный индекс характеристической моды с k-й меткой фокуса внимания; $I\left(M_{k+1}\left(t_{ex}\right)\right)$ - пространственный индекс характеристической моды с k+1-й меткой фокуса внимания; $\exists$ $e_{jl}$ – существование отношения пространственной связности j-й и l-й характеристических мод; $\mathrm{mod}(h'_{i\ell})\left\langle I\left(M_k\left(t_{ex}\right), M_{k+1}\left(t_{ex}\right)\right)\right\rangle$ – производная характеристическая мода $h'_{jl}$ с пространственным индексом, состоящим из меток $M_k(t_{ex})$ и $M_{k+1}(t_{ex})$.

Графически связность этих непроизводных характеристических мод через сформированную производную характеристическую моду $h'_{jl}$ в концепте производного структурного нейрона-детектора можно представить в виде следующего графа (рис. 11).

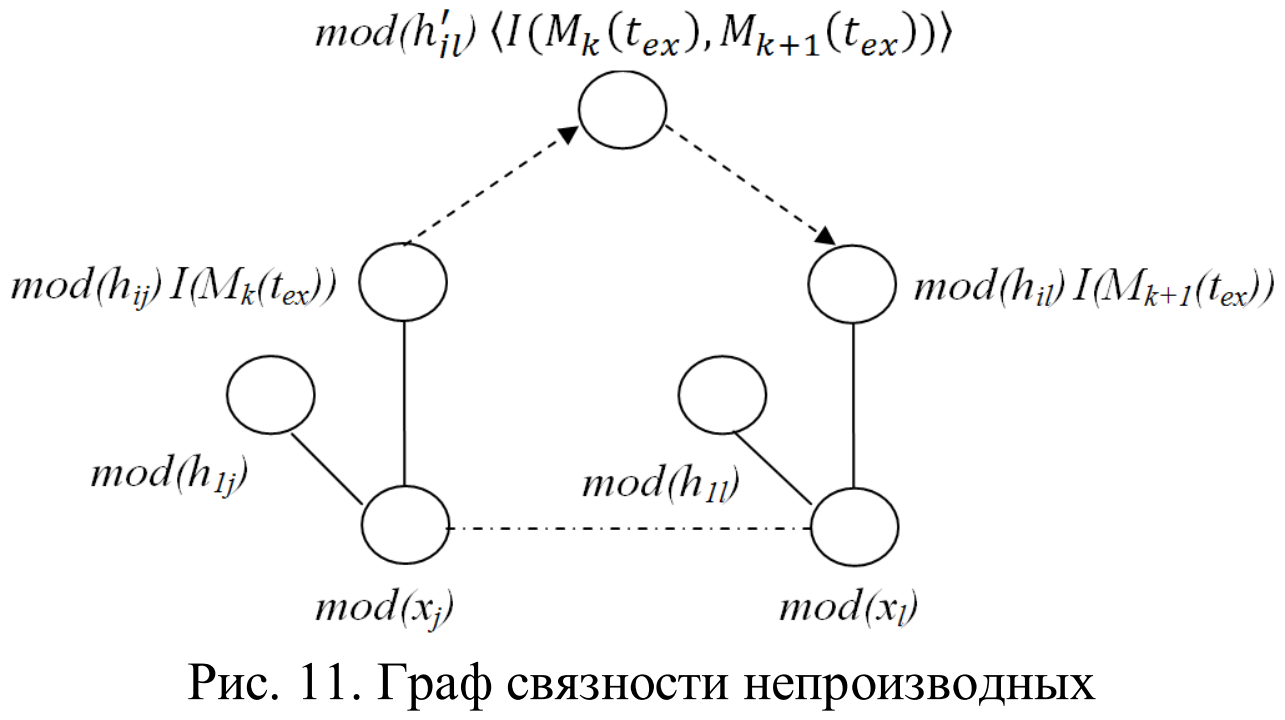

Рис. 11. Граф связности непроизводных характеристических мод mod ($h_{ij}$) и mod ($h_{il}$), соответствующих структурных мод mod ($x_j$) и mod ($x_l$), а также производной характеристической моды mod ($h'_{jl}$) в структуре концепта

Связность непроизводных характеристических мод с производной модой является направленной и обозначена на рис. 11 в виде дуг, что определяет свойство антисимметричности в отношении порядка непроизводных характеристических мод, который, в свою очередь, определяется последовательностью меток фокуса внимания. Направленность этих дуг формирует сильную связность рассмотренных мод.

Производные характеристики, сформированные в результате сравнения непроизводных характеристик, могут быть:

- количественными, значения которых определяются модулями разности адресных составляющих реакций пресинаптических нейронов-детекторов непроизводных характеристик:

$$\mathrm{mod}(h'_{jl}) = f\left|y_j(a) - y_l(a)\right|, \quad (24)$$

где mod($h'_{jl}$) – значение количественной производной характеристической моды. Это значение определяется как результат детектирования (функция f) нейроном-детектором реакции нейрона-анализатора, осуществляющего сравнения значений адресных составляющих $y_j(a)$ и $y_l(a)$ реакций j-го и l-го пресинаптических нейронов-детекторов непроизводных характеристик, связных по меткам фокуса внимания, и определения модуля разности этих значений;

- качественными, значения которых определяются отношениями строгого порядка либо тождества адресных составляющих реакций пресинаптических нейронов-детекторов непроизводных характеристик:

$$\mathrm{mod}(h^{*'}_{jl}) = f(y_j(a) * y_l(a)), \quad (25)$$

где mod($h^{*}_{jl}$) – значение качественной производной характеристической моды. Это значение получено в результате детектирования (функция f) нейроном-детектором реакции нейрона-анализатора, осуществляющего сравнения значений адресных составляющих $y_j(a)$ и $y_l(a)$ реакций j-го и l-го пресинаптических нейронов-детекторов непроизводных характеристик, связных по меткам фокуса внимания, и установления отношения *: строгого порядка ">", "<" либо равенства "=" между этими значениями.

Таким образом, реакции $y_j(a)$ и $y_l(a)$ нейронов-детекторов непроизводных характеристик, связные по меткам фокуса внимания, одновременно поступают на входы нейронов-анализаторов двух типов: количественных и качественных производных характеристик.

Количественные отношения адресных составляющих $y_j(a)$ и $y_l(a)$ являются инвариантными относительно их конкретных значений, т.е. одному значению количественной производной характеристической моды mod($h'_{jl}$) может соответствовать множество значений непроизводных характеристик. Следовательно, для значения количественной производной моды в структуре концепта вторичного (производного) структурного нейрона-детектора уже не имеет значения порядок следования меток фокуса внимания в структуре связных с ней мод непроизводных характеристик. Т.е. метки фокуса внимания нужны для связности значений непроизводных характеристик на этапе формирования значения производной характеристики и на этапе установления связности производных мод с образованной непроизводной модой в структуре концепта вторичного структурного нейрона-детектора. Одна-ко когда связность данных мод в процессе обучения уже установлена и эти моды активированы входными сигналами, порядок следования меток фокуса внимания и даже сами метки уже не имеют значения. Это обстоятельство позволяет вообще отказаться от формирования значений производных характеристик после того как были сформированы в процессе обучения соответствующие моды концепта вторичного структурного нейрона-детектора.

Качественные отношения адресных составляющих $y_j(a)$ и $y_l(a)$ являются инвариантными относительно их количественного значения и, следовательно, качественные моды еще более устойчивы (менее подвержены модификации в процессе обучения) в концепте структурного нейрона-детектора, чем количественные моды. Т.е., значение одной качественной производной моды $mod(h^{*'}_{jl})$ будет определять связность пар множества значений количественных непроизводных мод, если между ними в процессе анализа будет установлено соответствующее отношение порядка. Это говорит о том, что именно качественные моды определяют наиболее существенные (наиболее глубокие или сильные) признаки образов, относящихся к одному классу распознавания. Тогда, ориентированность дуг связности производной характеристической моды $mod(h'_{jl})$ в структуре концепта на рис.11 имеет значение только для качественной моды $mod(h^{*'}_{jl})$, т.к. для количественной производной характеристической моды эта связность имеет свойство симметричности.

Как уже отмечалось, к качественным отношениям порядка относятся отношения типа: ">", "<", "=", либо другие типы отношений, которые характеризуют упорядоченность адресных составляющих нейронов-детекторов непроизводных характеристик на определенной качественной шкале. Для объектов КМ этими отношениями могут быть отношения между биполярными значениями непроизводных качественных характеристик, которые определяются в процессе активного перцептивного акта. Например, изменение направления ориентации отрезков при выбранном направлении обхода контура изображения "по часовой стрелке" либо "против часовой стрелки" формируют значения производных биполярных характеристик - выпуклость либо вогнутость элементов контура изображения и всего изображения (контура) в целом.

Данные качественные биполярные отношения являются либо симметричными, как например отношение равенства "=", либо зеркально симметричными (антисимметричными), как отношения строгого порядка ">" либо "<". Свойство зеркальной симметрии в отношениях строгого порядка может быть отражено следующим выражением:

$$\forall\, a,b:\ aRb = b\check{R}a, \qquad (26)$$

где a,b – значения сравниваемых параметров; R – отношение строгого порядка; $\check{R}$ – противоположное R - зеркальное отношение строгого порядка.

Это говорит о том, что для данных видов отношений не имеет значения порядок следования меток фокуса внимания, связывающих сравниваемые параметры, в данном случае – значения непроизводных характеристик. Тогда, после того как в процессе обучения структурного нейрона-детектора, сопровождающегося последовательным активным перцептивным актом, будет сформирована качественная производная характеристическая мода, определяющая ориентированное отношение связности непроизводных характеристических мод (рис.11), необходимости в детектировании значения данной моды в процессе последующего параллельного восприятия уже не будет.

Рассмотренный характер количественных и качественных производных мод в структуре концепта производного структурного нейрона-детектора говорит о том, что данные моды могут быть ассоциативными, т.е переходить в активное состояние в моменты параллельного восприятия (моменты распознавания хорошо знакомых образов, что не связано с процессом активного воспиятия) даже без формирования входных сигналов - значений производных характеристик.

Введем следующее определение.

*Определение 11.* Ассоциативной модой $Amod(h_{ij})$ концепта производного структурного нейрона-детектора называется количественная либо качественная производная характеристическая мода, для активизации - прямого возбуждения которой необходимо и достаточно только возбуждения связных с ней метками фокуса внимания, в процессе обучения и в моменты времени активного перцептивного акта, соответствующих непроизводных характеристических мод.

Так как $Amod(h_{ij})$ в процессе обучения становится инвариантной относительно конкретных значений связных с ней непроизводных характеристических мод, то эта мода может принадлежать концепту даже тогда, когда данные непроизводные моды будут выведены (значением весового коэффициента) из структуры концепта.

Итак, рассматриваемая мода будет переходить в активное состояние с прямым, а не ассоциативным возбуждением в моменты времени параллельного восприятия образа (который был ранее запомнен в процессе активного перцептивного акта) даже без поступления на вход данного структурного нейрона-детектора реакции соответствующего пресинаптического характеристического нейрона-детектора, формирующего значение этой производной характеристической моды. Необходимым и достаточным условием возбуждения ассоциативной моды $Amod(h_{ij})$ будет только существование во входном векторе сигналов данного нейрона-детектора соответствующих реакций детекторов количественных непроизводных характеристик ($y_j(a)$ и $y_l(a)$), которые имели связность $e_{jl}$ в процессе обучения и, возможно, даже вышли из структуры концепта:

$$\exists\,(y_j(a),y_l(a),e_{jl}) => \overline{\mathrm{A\,mod}\left(h_{ij}\right)}, \qquad (27)$$

где $\overline{\mathrm{A\,mod}\left(h_{ij}\right)}$ – возбужденная ассоциативная производная характеристическая мода.

Существование ассоциативных мод в концепте структурного нейрона-детектора гипотетически может служить информационной моделью активных свойств дендрита и объяснять нейропсихологическое отличие процессов параллельного восприятия уже знакомого образа и последовательного активного перцептивного акта при его первоначальном восприятии, т.е. в процессе обучения.

Так как ассоциативная мода в процессе обучения становится связной с множеством пар значений непроизводных характеристик, то и ее весовой коэффициент принадлежности концепту $w_{ik}$ также является связным с весовыми коэффициентами мод данных характеристик. Кроме того, одна ассоциативная мода может быть связана с множеством структурных мод. Это означает, что $w_{ik}$ ассоциативной моды меняется в каждый момент времени встречного обучения (при наличии управляющего сигнала обучения z) либо самообучения (при возбуждении нейрона-детектора класса распознавания на соответствующей карте детекторов) при изменении весовых коэффициентов любых пар значений мод непроизводных характеристик ее образующих. Множество этих пар значений мод образуют матрицу AW - память данной ассоциативной моды, по структуре аналогичную матрице весовых коэффициентов W.

Исходя из типов производных характеристик, формируемых нейронами-анализаторами, определим, что нейроны-анализаторы могут быть двух типов: количественными и качественными.

Реакция количественного нейрона-анализатора определяет модуль разности адресных составляющих реакций пресинаптических нейронов-детекторов непроизводных характеристик. Значение результата сравнения (модуль разности) формирует уровень возбуждения нейрона-анализатора – составляющую его реакции $y_i(b)$, которая играет роль не только в процессе α-конкуренции, но и несет основную информационную нагрузку в качестве "нейронного кода" нейрона-анализатора. Таким образом, нейрон-анализатор может демонстрировать в своей реакции множество уровней возбуждения в отличие от единственного уровня возбуждения конкретного нейрона-детектора производных характеристик. Это определяет универсальность нейрона-анализатора по обработке множества значений одной непроизводной характеристики и позволяет говорить о некоторой аналогии нейронов-анализаторов с градуальными, не спайковыми нейронами (предетекторами базисных признаков) сенсорной системы, в которых значение амплитудного либо частотного сигнала (реакции) зависит от силы раздражителя [34].

Реакция качественного нейрона-анализатора формируется одновременно и в паре с реакцией количественного нейрона-анализатора на основе сравнения одних и тех же значений реакций пресинаптических нейронов-детекторов непроизводных характеристик, связных по меткам фокуса внимания $M_i$ и $M_{i+1}$. Уровень возбуждения качественного нейрона-анализатора $y_i(b)$ определяет качественные отношения: равенства "=" либо строгого порядка "<", ">" между сравниваемыми величинами. Тогда, для изображений КМ составляющая $y_i(b)$ реакции качественного нейрона-анализатора может иметь всего три значения: $b_1$, $b_2$, $b_3$, которые будут соответствовать определенному типу отношений.

Информационная модель нейрона-анализатора имеет следующие особенности:

1. Архитектура связей нейрона-анализатора с пресинаптическими нейронами-детекторами непроизводных характеристик имеет топическую структуру, т.е. связь осуществляется в пределах определенных пространственных окон $W_i$. Тогда, пространственный индекс реакции нейрона-анализатора формируется системой внимания таким же образом, как и для нейронов-детекторов непроизводных характеристик и элементов.

2. Функционирование и обучение нейрона-анализатора осуществляется только в моменты активного перцептивного акта, т.е. под управлением сигнала z, который формируется системой внимания (предположим, произвольного). Тогда, концепт нейрона-анализатора $Con(an_i)$, где $an_i$ – i-й нейрон-анализатор an (от англ. analyzer – анализатор), также формируется в процессе активного перцептивного акта и представляет собой память множества пар адресов пресинаптических нейронов-детекторов связных по меткам фокуса внимания.

Таким образом, детекторный подход предполагает, что в процессе формирования значений характеристик структурных элементов и образа в целом осуществляется последовательное чередование основных информационных составляющих реакций нейроэлементов. Этот процесс представлен на рис. 12.

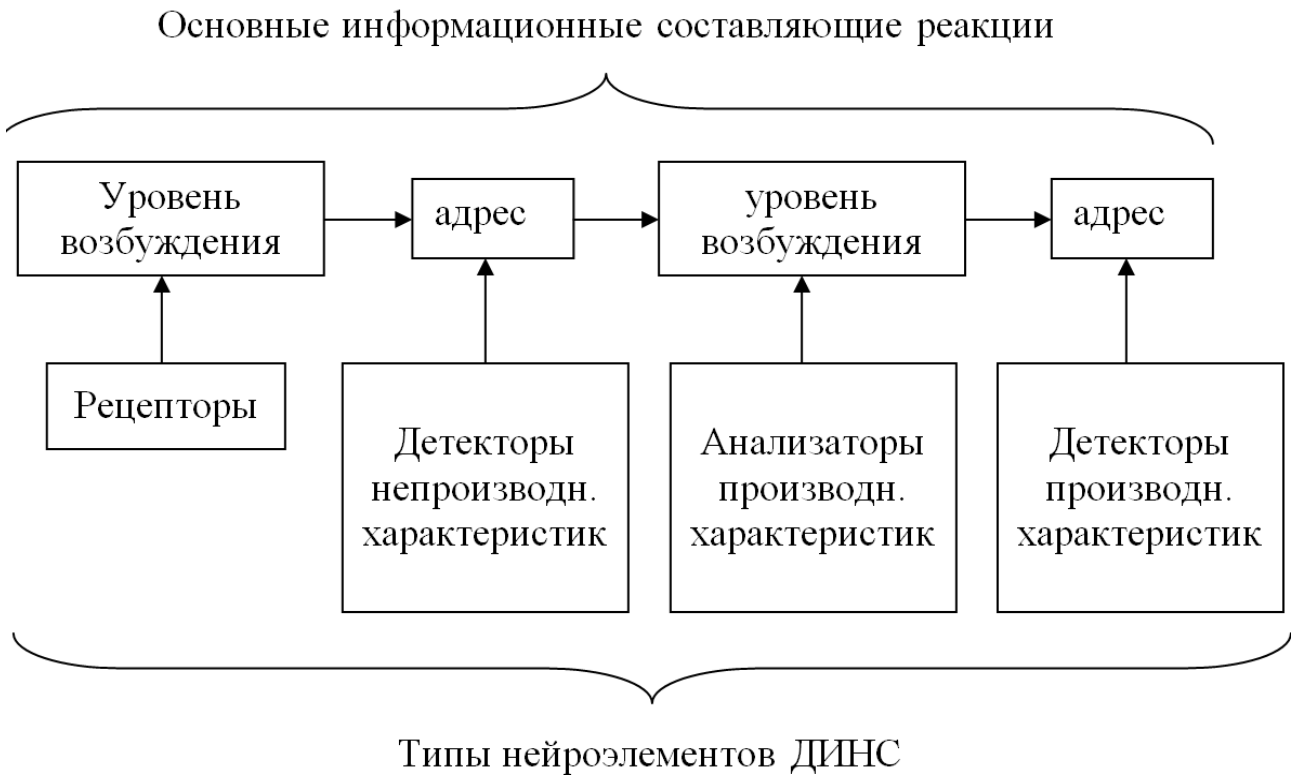

Рис.12. Чередование основных информационных составляющих реакций нейроэлементов ДИНС

**6. Архитектура ДИНС.** Особенностью архитектуры ДИНС, концептуально отличающей ее от архитектуры коннекционистских ИНС, является ее взаимосвязь с системой внимания, репрезентатив-ной системой, последующим (третичным) уровнем обработки визуальной информации и другими системами ДИНС. Обучение ДИНС заключается: в обучении не всех нейронов сети, а только отдельных нейронов-детекторов на вторичном (не топическом) уровне обработки информации, которое реализуется разработанной процедурой встречного обучения; а также в структуризации карт нейронов-детекторов в процессе самообучения ДИНС под управлением нейронов новизны. Нейрон новизны - нейрон нового типа, информационная модель которого также разработана в рамках детекторного подхода.

Под картой нейронов-детекторов будем понимать архитектурно заданное множество нейронов-детекторов одного уровня обработки информации, связанных латеральными тормозными связями, т.е. участвующими в процессе α-конкуренции. Все нейроны-детекторы (НД) одной карты детектируют образы, принадлежащие одному классу распознавания, например: одной буквы алфавита, либо одной цифре, либо одной геометрической фигуре и др. Структуризация карты заключается в определении значения и места каждого возбужденного НД в формируемой в процессе обучения онтологической структуре детекторов класса распознавания. Структура карты детекторов ДИНС соответствует современным представлениям нейропсихологии о топологии нейронов-детекторов во вторичных зрительных зонах, их возбуждении и взаимодействии в процессе опознания различных образов [35].

**6.1. Общая структура ДИНС**. Общая структура ДИНС для распознавания изображений КМ представлена на рис. 13, где изображены следующие модули и элементы:

$X_1$ - информация о дистальном стимуле X, поступающая в сенсорную оптическую подсистему A;

$X_2$ - информация о дистальном стимуле X, поступающая в сенсорную подсистему презентативной системы N. Этой системой может быт например, звуковая система, моделирующая слуховое восприятие и распознавание;

A - модуль ДИНС, образующий сенсорную оптическую подсистему ДИНС, которая моделирует сенсорную систему зрительного восприятия. Данная подсистема состоит из сенсоров (рецепторов), сгруппированных в сенсорные (рецептивные) поля, имеющие структуру прямоугольных матриц. Эти матрицы образуют окна восприятия $W_i$, которые имеют различную размерность;

I - первичный (топический) уровень обработки визуальной информации на котором осуществляется: детектирование непроизводных структурных элементов и их характеристик в подсистеме "Что", а также детектирование пространственных характеристик (формирование пространственных индексов I в структуре реакций нейронов ДИНС) в подсистеме "Где";

II - вторичный (не топический) уровень обработки визуальной информации на котором осуществляется: формирование производных характеристик распознаваемого образа и его структурных элементов в результате анализа (сравнения) значений непроизводных характеристик, осуществляемого нейронами-анализаторами; детектирование значений производных характеристик соответствующими нейронами-детекторами; детектирование (распознавание) сложных структур - цельных образов;

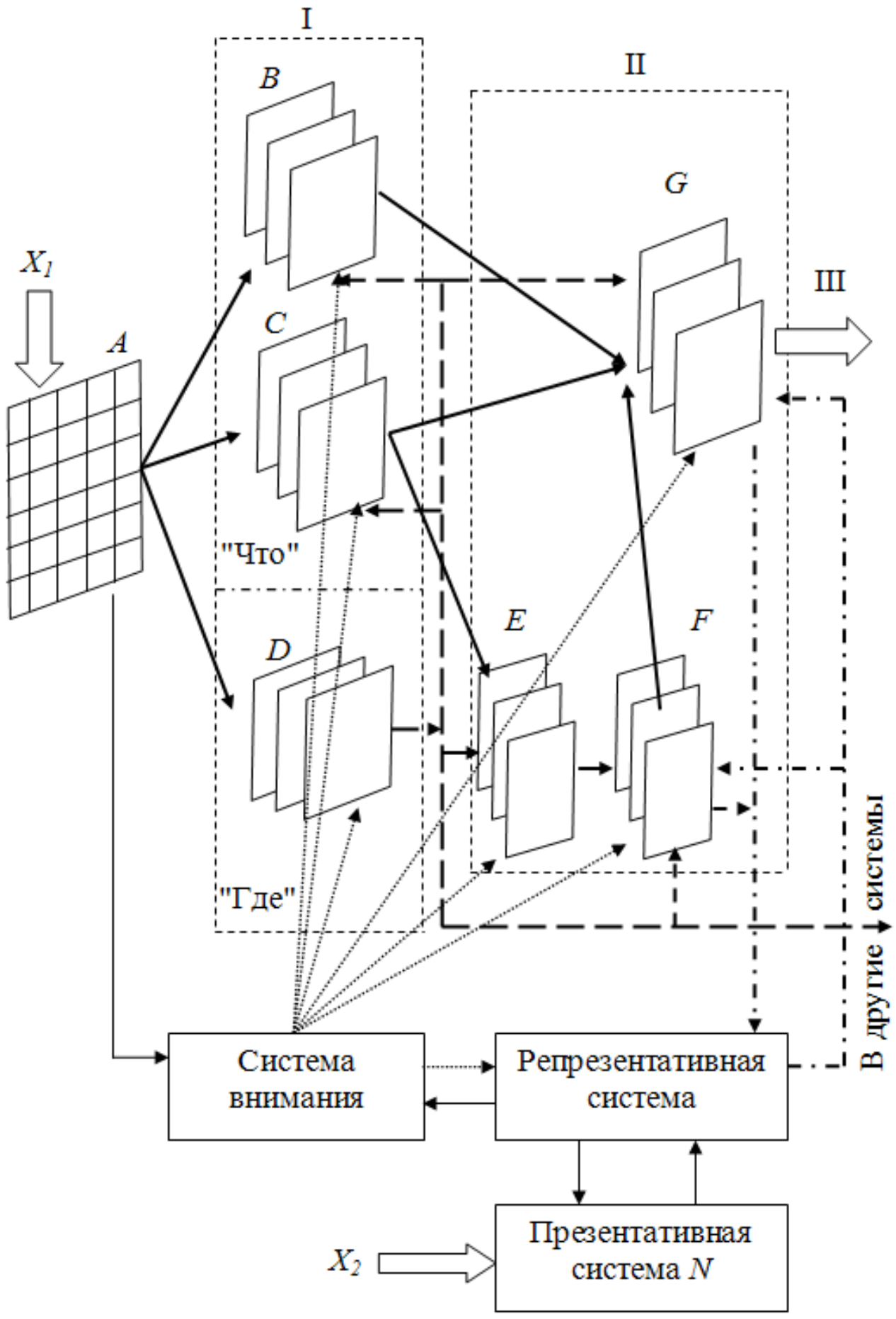

Рис. 13. Общая структура ДИНС

III - третичный (не топический) уровень обработки визуальной информации на котором осуществляется распознавание связанных образов - сцен;

Система внимания. В данной структуре ДИНС система внимания осуществляет моделирование активного перцептивного акта: формирование меток фокуса внимания. Эта система работает под управлением репрезентативной системы ДИНС;

Репрезентативная система (RpS) - система ДИНС, осуществляющая на II уровне обработки информации реализацию процесса встречного обучения нейронов-детекторов;

Презентативная система N - презентативная система (PrS) ДИНС, осуществляющая процесс восприятия и распознавания того же дистального стимула X, но уже с другой сенсорной информацией $X_2$. Примером подобной системы может служить звуковая система. Эта система, в данном случае, принимает участие в процессе встречного обучения;

B - модуль ДИНС, содержащий множество групп карт (плоскостей) нейронов-детекторов непроизводных структурных элементов;

C - модуль ДИНС, содержащий множество групп карт (плоскостей) нейронов-детекторов непроизводных характеристик структурных элементов. Модули B и C образуют подсистему "Что" ДИНС;

D - модуль ДИНС, содержащий множество групп карт (плоскостей) нейронов-детекторов пространственного положения стимулов в рецепторном поле восприятия A. Данный модуль образует подсистему "Где" ДИНС;

E - модуль ДИНС, содержащий множество групп карт (плоскостей) нейронов-анализаторов, формирующих значения производных характеристик на вторичном уровне обработки визуальной информации;

F - модуль ДИНС, содержащий множество групп карт (плоскостей) нейронов-детекторов значений производных характеристик на вторичном уровне обработки визуальной информации;

G - модуль ДИНС, содержащий множество групп карт (плоскостей) нейронов-детекторов производных структур (цельных образов) на вторичном уровне обработки визуальной информации.

Представленная на рис. 13 структура ДИНС функционирует следующим образом.

1. В модуль A и презентативную систему N одновременно поступает информация, соответственно $X_1$ и $X_2$, о дистальном стимуле X, подлежащем распознаванию. Например, $X_1$ - зрительная информация (зрительный образ), $X_2$_- звуковая информация (семантический определитель или "имя" образа). Информация $X_1$ воспринимается сенсорами модуля A и параллельно передается на нейроны-детекторы модулей B, C, D первичного уровня обработки информации, а также в систему внимания для реализации процесса активной перцепции. Особенностью данного этапа является топическая (ретинотопическая) структура связей сенсоров модуля A и нейронов-детекторов модулей B, C, D. Информация $X_2$ воспринимается сенсорами PrS N в которой осуществляется аналогичный процесс обработки информации.

2. В модулях B, C и D параллельно (синхронно) в результате α-конкуренции возбуждается группа (ансамбль) нейронов-детекторов, функции и местоположение которых заданы архитектурно (в процессе проектирования). При формировании реакций нейронов-детекторов модулей B и C происходит добавление (инкапсуляция) пространственной составляющей - пространственного индекса I. Значение пространственного индекса I формируется нейронами-детекторами модуля D.

3. В случае осуществления в ДИНС под управлением системы внимания активного перцептивного акта, система внимания формирует метки фокуса внимания, которые инкапсулируются в структуру пространственного индекса I. Система внимания программирует процесс активного перцептивного акта.

4. Реакции нейронов-детекторов модулей B и C поступают параллельно (синхронно) в процессе пассивной перцепции, либо параллельно-последовательно в процессе активного перцептивного акта, на входы нейронов-детекторов модуля G вторичного уровня обработки информации. Кроме того, в процессе активной перцепции реакции нейронов-детекторов модуля C также поступают и на входы нейронов-анализаторов модуля E. В свою очередь, реакции нейронов-анализаторов поступают на входы нейронов-детекторов модуля F, которые формируют реакции, поступающие далее на входы нейронов-детекторов модуля G. Параллельно-последовательный процесс формирования реакций нейронов на вторичном уровне обработки информации осуществляется в период времени экспозиции $t_{ex}$, который определяется системой внимания. Отличительной особенностью процесса взаимосвязи нейронов первичного и вторичного уровней обработки информации, а также взаимосвязи между нейронами на вторичном уровне, является не топический характер связей. В ДИНС эти связи реализуются по архитектуре "общая шина".

5. На вторичном уровне обработки информации реализуется процесс встречного обучения нейронов-детекторов презентативной и репрезентативной систем. Этот процесс заключается в формировании концептов нейронов-детекторов, а также в формировании структуры (топологии) карт (плоскостей) на которых они расположены.

6. Реакции нейронов-детекторов модуля G передаются на входы нейронов третичного уровня обработки информации III, который, как и модели других систем мозга, в данной работе не рассматривается.

**6.2. Архитектура ДИНС на не топическом (вторичном) уровне обработки информации.** Именно модуль G содержит структурные нейроны-детекторы, информационные модели и процесс обучения которых были рассмотрены в этой работе. Архитектура данного модуля формируется под управлением нейрона новизны (НН) в процессе встречного обучения и самообучения его НД. Очевидно, что функции НН ДИНС отличаются от функций биологических нейронов новизны. Функции НН ДИНС определены необходимостью реализации детекторной парадигмы построения ИНС, основанной на принципе абсолютной адресации нейронов-детекторов.

Каждый нейрон новизны управляет процессом обучения и структуризации своей карты нейронов-детекторов (рис. 14). Каждая подобная карта реагирует только на цельные образы, принадлежащие одному классу распознавания, например: конкретной геометрической фигуры, отдельному символу алфавита, лицу человека (в том числе лицу конкретного человека) и т.д. Таким образом, количество карт НД - кластеров соответствует количеству классов распознавания образов. Каждый НН и, соответственно, каждая карта НД связывается определенным сигналом встречного обучения z, поступающим из RpS. Для карты нейронов-детекторов модуля G, изображенной на рис.14, это сигнал $z_1$.

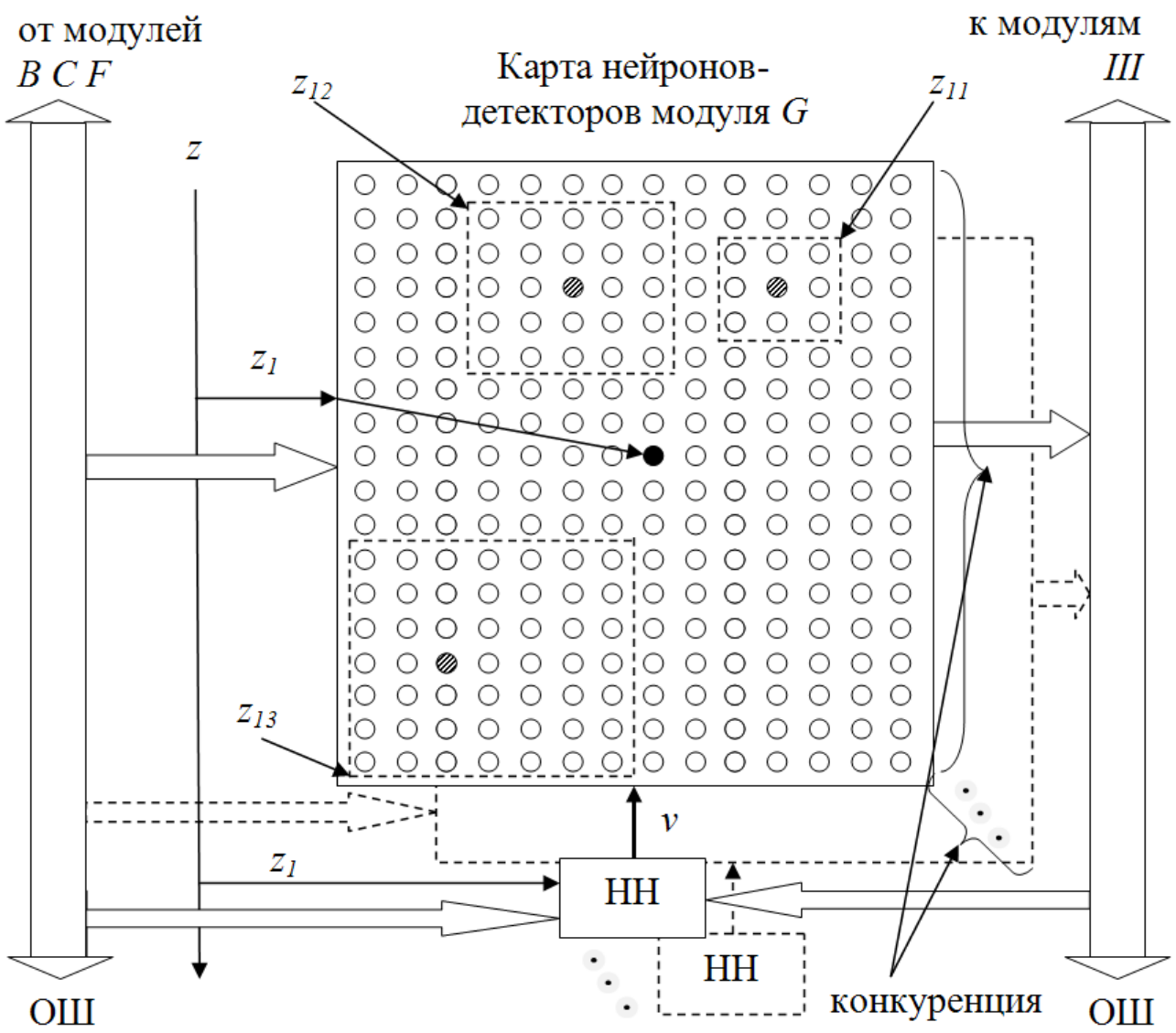

Рис. 14. Архитектура карты нейронов-детекторов модуля G

Рассмотрим процесс формирования архитектуры одной карты нейронов-детекторов модуля G и моменты функционирования нейрона новизны, управляющего процессом структуризацией данной карты. Данный процесс уточняет процедуры встречного обучения и самообучения.

1. Если в карте НД нет ни одного обученного нейрона-детектора. В этом случае при поступлении первого сигнала обучения $z_1$ синхронно с входным вектором сигналов X в период времени экспозиции образа $t_{ex}$, НН запоминает сигнал $z_1$ в своем блоке памяти, т.е. происходит "захват" НН, его связывание с соответствующим нейроном-детектором RpS ДИНС. После этого, НН генерирует сигнал "захвата" v нейрона-детектора своей карты. Данный сигнал содержит адрес нейрона-детектора подлежащего "захвату". Первым "захваченным" нейроном-детектором будет центральный детектор карты (на рис.14 изображен черной точкой). Этот нейрон запоминает входной вектор X (X=Con) и сигнал обучения $z_1$ (происходит связывание данного НД с соответствующим НД RpS ДИНС). В дальнейшем именно этот нейрон-детектор будет подвергаться обучению (будет корректироваться его концепт) и он станет нейроном-детектором класса распознавания или НД кластера (от англ. cluster - скопление, рой). Определение адресов НД, подлежащих "захвату" происходит неслучайным образом. Этим процессом управляет блок захвата в информационной модели нейрона новизны.

2. Если в карте уже есть обученные НД и в процессе встречного обучения в период времени экспозиции образа $t_{ex}$ параллельно с входным вектором сигналов X приходит сигнал $z_{1i}$, который принадлежит классу сигналов обучения $z_1$ (что определяется по значению адресной составляющей сигнала $z_{1i}$) но не совпадает с ранее запомненным, и обученные НД данной карты не реагируют (несовпадение X с Con), то НН также запоминает этот сигнал в блоке Z и, генерируя соответствующий сигнал v, "захватывает" свободный нейрон-детектор карты, который будет обучаться и является центром определенной свободной зоны карты - подкластера (на рис. 14 эти нейроны, связанные сигналами $z_{11}$, $z_{12}$, $z_{13}$, изображены заштрихованными кружками, а зона подкластера выделена пунктирной линией). Данные нейроны будут детекторами подкласса распознавания образов и обучаться также, как НД кластера (см. п.1).

3. Если в карте уже есть обученный НД и в период времени экспозиции образа $t_{ex}$ параллельно с ранее запомненным сигналом $z_{1i}$ приходит входной вектор X, связные структурные моды которого соответствуют концепту Con данного НД, а характеристические моды отличаются от ранее запомненных, то данный НД возбуждается и происходит его встречное обучение, заключающееся в корректировке его концепта Con. Кроме того, НН сравнивает суммы водных сигналов $x_i \in X$ с актуальной реакцией (уровнем возбуждения y(b)) возбужденного НД (см. выражение (28)) и, в случае $\sum_{i=1}^{k} x_i > y(b)$, "захватывает" ближайший к возбужденному НД свободный нейрон. "Захваченный" НД будет детектором примера распознавания с Con, соответствующим входному вектору X. Данный НД не будет связан с сигналом $z_{1i}$ и его концепт, в дальнейшем, не будет модифицироваться. Очевидно, что в этом случае количество НД примеров распознавания будет велико. Для оптимизации структуры карты необходимо удалить те НД примеров распознавания, которые были возбуждены только один раз или были возбуждены достаточно давно. Для этой цели необходимо использовать механизм "забывания" [29].

4. Если в карте уже есть обученные НД и в период времени экспозиции образа $t_{ex}$ параллельно с ранее запомненным сигналом $z_{1i}$ приходит входной вектор X, но обученные НД данной карты не реагируют (несовпадение X с Con), то НН "захватывает" ближайший свободный нейрон к НД кластера (подкластера), связного данным сигналом обучения. "Захваченный" НД будет альтернативным детектором данного класса (подкласса) распознавания, который также будет связан сигналом $z_{1i}$ и его концепт также может модифицироваться в процессе встречного обучения.

5. Если в карте уже есть обученные НД и сигнал обучения $z_1$ не поступает, но поступает входной вектор сигналов X, на который реагирует хотя бы один детектор карты, то НН осуществляет сравнение суммы водных сигналов $x_i \in X$ с актуальной реакцией кластера - уровнем возбуждения y(b) победившего НД:

$$\sum_{i=1}^{k} x_i * y(b) \,, \qquad (28)$$

где * - оператор сравнения.

Возбуждение хотя бы одного НД карты свидетельствует о принадлежности входного стимула классу распознавания данной карты. В этом случае возможны три результата сравнения:

a) $\sum_{i=1}^{k} x_i > y(b)$;

b) $\sum_{i=1}^{k} x_i < y(b)$;

c) $\sum_{i=1}^{k} x_i = y(b)$.

В случае a) происходит самообучение ДИНС, заключающееся в "захвате" НН нового нейрона-детектора примера класса распознавания - ближайшего свободного нейрона к возбужденному НД. При самообучении не происходит корректировка концептов, а только структуризация карты детекторов.

В случае b) происходит распознавание входного вектора сигналов с учетом возбуждения ассоциативных мод концепта НД либо ассоциативное возбуждение НД при неполном входном векторе X.

В случае c) происходит прямое распознавание входного вектора сигналов.

Таким образом, корректировка концептов НД кластера или подкластера осуществляется только в процессе встречного обучения при наличии управляющего сигнала $z_{i.}$, а при его отсутствии, при условии $\sum_{i=1}^{k} x_i > y(b)$, происходит самообучение ДИНС, заключающееся в "захвате" и обучении НД примера класса распознавания. Концепт НД примера класса распознавания оказывается несвязанным с сигналом обучения $z_{i.}$, что не позволяет осуществить его встречное обучение.

6. Если в карте уже есть обученный НД и поступает ранее запомненный и связанный с данным НД сигнал обучения $z_1$, но входной вектор сигналов X не поступает, то происходит ассоциативное возбуждение данного НД. НН не участвует в данном процессе.

## Заключение

Таким образом, детекторный принцип построения ИНС, являясь альтернативой коннекционистской парадигме, позволяет комплексно решить большинство указанных проблем данной парадигмы и построить биологически мотивированную ДИНС, которая также обладает рядом технических преимуществ по сравнению с современными коннекционистскими ИНС.

Для практической проверки основных положений, методов и процедур детекторного принципа построения ДИНС и определения технических характеристик сети было осуществлено построение программной модели ДИНС (карт детекторов ее вторичного уровня) и проверка ее эффективности при распознавании изображений рукописных цифр базы MNIST [36]. Выбор данной базы изображений был обусловлен тем, что основные положения детекторного принципа рассматривались применительно к распознаванию объектов "Контурного мира" к которым могут быть отнесены и рукописные цифры. Кроме того, данная библиотека традиционно используется для анализа и сравнения различных ИНС.

Символы библиотеки были подвергнуты предварительной обработке: скелетизации отрезками прямых с выделением структурных критических точек и векторизации отрезков с определением точек "захвата" контура и направлений обхода контура изображения. В ходе предварительной обработки и имитации встречного обучения ДИНС была реализована модель активного перцептивного акта, а в ходе тестирования - модель параллельного распознавания. Результаты программного моделирования на языке Python показали, что для обучения нейронов-детекторов вторичного уровня обработки ДИНС достаточно только одной эпохи обучения, при этом количество ошибок распознавания на тестовой выборке из 10000 символов составляет 34 (0,34%), что сопоставимо с одними из лучших результатов распознавания символов библиотеки, продемонстрированных 6-уровневой сетью (784-40-80-500-1000-2000-10) и глубокой сверточной сетью (1-20-40-60-80-100-120-120-10) [36]. При этом общее количество возбужденных нейронов-детекторов вторичного уровня по всем 10 картам составило 249 детекторов.

## Список литературы

## ДЕТЕКТОРНИЙ ПРИНЦИП ПОБУДОВИ ШТУЧНИХ НЕЙРОННИХ МЕРЕЖ ЯК АЛЬТЕРНАТИВА КОННЕКЦІОНІСТСЬКІЙ ПАРАДИГМІ

Ю.В. Паржин

*Штучні нейронні мережі (ШНМ) є неадекватними біологічним нейронним мережам. Ця неадекватність проявляється у використанні застарілої моделі нейрону та коннекціоністській парадигмі побудови ШНМ. Результатом даної неадекватності є існування множини недоліків ШНМ та проблем їхньої практичної реалізації. В статті пропонується альтернативний принцип побудови ШНМ. Цей принцип отримав назву детекторного принципу. Основою детекторного принципу є розгляд властивості зв'язності вхідних сигналів нейрону. У даному принципі використовується нова модель нейрону-детектору, новий підхід до навчання ШНМ - зустрічне навчання та новий підхід до формування архітектури ШНМ.*

## DETECTIVE PRINCIPLE OF ARTIFICIAL NEURAL NETWORKS CONSTRUCTION AS ALTERNATIVE OF THE CONNECTIONIST PARADIGME

Yu.V. Parzhin

*Artificial neural networks (ANN) are inadequate to biological neural networks. This inadequacy is manifested in the use of the obsolete model of the neuron and the connectionist paradigm of constructing ANN. The result of this inadequacy is the existence of many shortcomings of the ANN and the problems of their practical implementation. The alternative principle of ANN construction is proposed in the article. This principle was called the detector principle. The basis of the detector principle is the consideration of the binding property of the input signals of a neuron. A new model of the neuron-detector, a new approach to teaching ANN - counter training and a new approach to the formation of the ANN architecture are used in this principle.*